\documentclass[10pt]{article}
\usepackage{blindtext}
\usepackage[a4paper, total={6in, 8in}]{geometry}
\usepackage[numbers]{natbib}
\usepackage{lastpage}
\usepackage[utf8]{inputenc} 
\usepackage[T1]{fontenc}    
\usepackage{hyperref}       
\usepackage{url}            
\usepackage{booktabs}       
\usepackage{amsfonts}       
\usepackage{nicefrac}       
\usepackage{microtype}      
\usepackage{xcolor}         
\usepackage{color,soul}
\usepackage{threeparttable}
\usepackage{array,booktabs} 
\usepackage{float}
\usepackage{adjustbox}
\usepackage{longtable}
\usepackage{microtype}
\usepackage{graphicx}
\usepackage[ruled,vlined,nofillcomment]{algorithm2e}
\usepackage{mathtools,amsmath,amsthm,amssymb,amsfonts,bm}
\usepackage{colortbl}
\usepackage{multirow}
\usepackage{tikz}
\usepackage{subcaption}
\usepackage{pgfplots}
\usepackage{enumitem}
\pgfplotsset{compat=newest}
\pgfplotsset{plot coordinates/math parser=false}
\usetikzlibrary{shapes,arrows,fit,calc,positioning,matrix}
\usetikzlibrary{shapes.geometric}
\usetikzlibrary{spy}
\tikzset{internal/.style={draw, ellipse, x radius=3cm, y radius=3cm, thick, text centered}}
\tikzset{subtree/.style={isosceles triangle, isosceles triangle apex angle=80, draw, shape border rotate=90, thick, text centered,minimum height=1cm}}
\tikzset{leaf/.style={draw, rectangle, thick, text centered, minimum height=0.5cm, minimum width=1cm}}
\tikzset{line/.style={draw, thick, -latex'}}
\tikzset{state/.style={draw, circle, fill=white,draw=black,thick,text centered}}
\tikzset{cell/.style={rectangle, thick, draw}}

\hypersetup{
    colorlinks=true,       
    linkcolor=black,          
    citecolor=black,        
    urlcolor=blue           
}

\newcommand{\T}{\mathbb{T}} 
\newcommand{\Lf}{\mathcal{L}} 
\newcommand{\I}{\mathcal{I}} 

\newtheoremstyle{break}{\topsep}{\topsep}{\itshape}{}{\bfseries}{}{\newline}{}
\theoremstyle{break}
\newtheorem{assumption}{Assumption}[section]
\newtheorem{theorem}{Theorem}[section]

\newtheorem{lemma}{Lemma}[section]
\newtheorem{definition}{Definition}[section]
\newtheorem{notation}{Notation}[section]

\begin{document}

\title{Divide, Conquer, Combine Bayesian Decision Tree Sampling}
\author{ Jodie A.~Cochrane \\\small \texttt{Jodie.Cochrane@newcastle.edu.au} 
   \and
  Adrian Wills \\ \small\texttt{Adrian.Wills@newcastle.edu.au } 
   \and
   Sarah J.~Johnson \\ \small\texttt{Sarah.Johnson@newcastle.edu.au} 
}

\date{}

\maketitle

\begin{abstract}

Decision trees are commonly used predictive models due to their flexibility and interpretability. This paper is directed at quantifying the uncertainty of decision tree predictions by employing a Bayesian inference approach. This is challenging because these approaches need to explore both the tree structure space and the space of decision parameters associated with each tree structure. This has been handled by using Markov Chain Monte Carlo (MCMC) methods, where a Markov Chain is constructed to provide samples from the desired Bayesian estimate. Importantly, the structure and the decision parameters are tightly coupled; small changes in the tree structure can demand vastly different decision parameters to provide accurate predictions. A challenge for existing MCMC approaches is proposing joint changes in both the tree structure and the decision parameters that result in efficient sampling. This paper takes a different approach, where each distinct tree structure is associated with a unique set of decision parameters. The proposed approach, entitled DCC-Tree, is inspired by the work in \citet{zhou2020divide} for probabilistic programs and \citet{cochrane2023rjhmctree} for Hamiltonian Monte Carlo (HMC) based sampling for decision trees. Results show that DCC-Tree performs comparably to other HMC-based methods and better than existing Bayesian tree methods while improving on consistency and reducing the per-proposal complexity. 

\end{abstract}

\section{Introduction}

The decision tree model is used extensively throughout various industries as it provides both the flexibility to describe data relations and the interpretability of the resultant model. Decision trees define a set of hierarchical splits that partition the input space into a union of disjoint subspaces. The datapoints associated with each subspace are then assumed to originate from the same distribution, which then defines the model predictions. Some popular methods to learn a decision tree are CART \citep{breiman1984classification}, ID3 \citep{quinlan1986induction}, and C4.5 \citep{quinlan1993c4}, each of which gives the output predicted by the learnt tree model as a point-estimate. 

It is important for any decision-making process to consider the uncertainty associated with a point-estimate prediction. As such, there has recently been growing emphasis on estimating the prediction uncertainty for machine learning models \cite{ghahramani2015probabilistic}. 
Incorporating uncertainty into predictions can be achieved in a mathematically coherent way by using Bayesian inference techniques. Under the Bayesian paradigm, probability distributions are used to define anything about the model that is not known for certain, including the model parameters themselves. The uncertainty in the model parameters induces uncertainty in the resulting predictions, which can be used to estimate confidence intervals via evaluation of an expectation integral. 

A current challenge for Bayesian decision trees is that a change in the tree structure fundamentally changes the meaning of the parameter vector. This issue manifests in the coupling between the tree topology and the associated decision parameters.
Evaluating the expectation integral therefore requires consideration of both the nonlinearity of the model and the association between the parameter vector and the tree topology. A common approach is to approximate the integral based on useful samples from the parameter distribution using a method known as Monte Carlo integration. The problem then shifts to how to generate these useful samples for the decision tree model.

To date, the literature has mainly focused on Markov chain Monte Carlo (MCMC) approaches to address this problem and attempt to explore the posterior distribution of the decision tree parameters. Random-walk MCMC was first proposed by \citet{chipman1998bayesian} and \citet{denison1998bayesian} where a set of local, tree-inspired proposals are used to move around the space. Over time, improvements have been made to that original set of proposal methods, but have remained based on random-walk methods \cite{wu2007bayesian,gramacy2008bayesian,pratola2016efficient}. 

Separately, there have been attempts to use Sequential Monte Carlo (SMC) sampling methods to explore the posterior distribution \cite{taddy2011dynamic,lakshminarayanan2013top}. In this case, many `particles' or trees are grown simultaneously, keeping better-performing trees (based on likelihood weights) at subsequent iterations, until a stopping criterion has been reached. This helps explore different areas of the posterior distribution by applying local, random perturbations to each tree. However, these transitions are similar to the localised, random-walk proposals of MCMC-based methods, although \citet{lakshminarayanan2013top} does provide a locally optimal proposal option to try and improve on the random-walk issues.

The recent work of \citet{cochrane2023rjhmctree} has shown the efficacy of using Hamiltonian Monte Carlo (HMC) within the MCMC framework to explore the decision tree space, where the decision tree structure has been softened to enable full use of the HMC benefits while remaining interpretable. HMC is a more efficient sampling method that uses gradient information from the likelihood to generate subsequent samples \cite{neal2011mcmc,betancourt2017conceptual}. Although the benefits of using HMC are clear, the authors note that the main impediment to their method is the number of intermediate HMC samples required at every iteration, drastically increasing computational overhead for each proposal.

In this paper, we investigate a different perspective by which to explore the posterior distribution of decision trees in an attempt to rectify the computational overhead while retaining the observed sampling efficiency of HMC.  Inspired by the work of \citet{zhou2020divide}, which proposes the Divide, Conquer, Combine (DCC) inference framework in the context of probabilistic single-line programs, the parameter space of the decision tree is described by an all-encompassing parameter vector with constant dimension. This is in contrast to previous Bayesian decision tree algorithms, where the dimension of the parameter vector varied throughout the sampling routine. Parameters corresponding to all possible tree topologies are included in the parameter vector, with a subset of this vector considered at each iteration. This concept forms the basis of the novel DCC-Tree sampling method presented in this paper.

This paper is divided into the following sections. Section 2 first presents a background on decision trees and Bayesian inference. 
The DCC-Tree algorithm is then discussed in Section 3, including details on the local approximation strategy and exploration of the different tree structures. This is followed by the presentation of the final algorithm. Section 4 then analyses the DCC-Tree algorithm and compares it to existing methods Bayesian decision tree methods. Section 5 concludes the paper with a discussion and final remarks.

\section{Decision Tree Model and Problem Formulation}
A standard binary decision tree is parameterised by
\begin{equation}
    \T = (\mathcal{T},\bm{\kappa},\bm{\tau},\bm{\theta}),
\end{equation}
where: 
\begin{itemize}
    \item $\mathcal{T} = (\mathcal{V},\mathcal{E})$ denotes the tree topology, which encompasses information about the set of tree nodes $\mathcal{V}$ and set of interconnections (edges) $\mathcal{E}$. Note that $\mathcal{V}$ contains both the set of internal nodes $\eta_j \in \I, j=1,\dots,n$ and the set of leaf nodes $\eta_{\ell_k} \in \Lf, k=1,\dots,n_\ell$.
    \item $\bm{\kappa} = \left[ \kappa_1,\kappa_2,\dots,\kappa_n \right]$ denotes the splitting indices of each internal node $\eta_j \in \I$,
    \item $\bm{\tau}= \left[ \tau_1,\tau_2,\dots,\tau_n \right]$ represent the splitting thresholds for each internal node $\eta_j \in \I$,
    \item $\bm{\theta} = \left[ \theta_1,\theta_2,\dots,\theta_{n_\ell} \right]$ represents the leaf node parameters for $\eta_{\ell_k} \in \Lf$.  
\end{itemize}

We refer to the dataset as $\mathcal{D} = \{ (\mathbf{x}_i, \mathbf{y}_i )\}_{i=1}^N$ which is comprised of $N$ sets of inputs $\mathbf{x}_i \in \mathcal{X}$ and outputs $\mathbf{y}_i \in \mathcal{Y}$. The input and output dimensionalities are denoted $n_x$ and $n_y$ respectively. In this paper, we will consider both regression and classification problems where the output space dimension is $n_y=1$, that is $\mathcal{Y} = \mathbb{R}$ or $\mathbb{Z}$, and the input space is real-valued $\mathcal{X} = \mathbb{R}^{n_x}$.

The traditional approach to constructing a decision tree is via a greedy one-step-ahead heuristic.
Predictions for a given input are made by traversing the tree, starting at the root node and recursively moving to one of the children nodes corresponding to the value of the relevant splitting indices for the given input. The traversal stops once a leaf node has been reached. The output is then related to the datapoints that are in that leaf node. In a standard decision tree, this corresponds to a single output value, such as the most represented class for classification trees, or the average value of all datapoints in that leaf for regression trees. For Bayesian decision trees, as will be discussed later, the output is assumed to be associated with the probability distribution at that leaf node. 

Bayesian inference offers a means to update the initial (or prior) probability of an event using information (or data) about the event. 
Probability theory provides a mathematically coherent way in which to express uncertainty about certain values or events. Bayes' Theorem can be used to combine the prior distribution with information about the system via the likelihood distribution to produce an updated posterior distribution. For prediction modelling, the distribution of interest is the posterior predictive distribution, which can be related to the posterior distribution on the model parameters.

The posterior predictive distribution can be used to quantify the uncertainty of model predictions for a new datapoint $(\mathbf{x}^*,\mathbf{y}^*)$ via
\begin{equation}\label{eq:bdt-pred}
    p(\mathbf{y}^*\mid \mathbf{x}^*,\mathcal{D}) = \int p(\mathbf{y}^*\mid \mathbf{x}^*,\Theta)p(\Theta\mid \mathcal{D})d\Theta
\end{equation} 
where $p(\mathbf{y}^*\mid \mathbf{x}^*,\Theta)$ refers to the distribution of the model output given a new input, and $p(\Theta\mid \mathcal{D})$ is the posterior on model parameters. 
In the case that this integral is not analytically tractable (as is the case for the decision tree model), one such method to evaluate this is via Monte Carlo integration, which is defined as follows, 
\begin{equation}\label{eq:predictive-dist-trees}
    p(\mathbf{y}^*\mid \mathbf{x}^*,\mathcal{D}) \approx \frac{1}{M}\sum_{m=1}^M p\left(\mathbf{y}^*\mid \mathbf{x}^*, \Theta^{i}\right), \quad  \Theta^{i} \sim p(\Theta\mid \mathcal{D})
\end{equation}
where a sample drawn from the posterior distribution on parameters is then used to compute the output of the corresponding model, which generates part of the Monte Carlo estimate. The problem now shifts to generating useful samples from the posterior distribution on parameters $\Theta^i \sim p(\Theta\mid \mathcal{D})$. Using Bayes Theorem, this distribution can be expressed as,
\begin{equation}\label{eq:bdt-bayes}
    p(\Theta \mid  \mathcal{D}) \propto p(\mathcal{D}\mid \Theta) p(\Theta)
\end{equation}
where $p(\mathcal{D}\mid \Theta)$ is the likelihood distribution of the observed data given the model parameters and $p(\Theta)$ denotes the prior distribution on parameters. 
Therefore, samples can be generated from the posterior distribution on parameters by using appropriate definitions for both the prior and likelihood distributions.

In this paper, we consider the situation in which the model is assumed to be a decision tree and we wish to determine the uncertainty on the decision tree parameters given a dataset. Although relatively easy to use to make point-wise predictions, defining uncertainty for the predictions from a decision tree model is a more difficult task. Once a set of parameters is selected, it is straightforward to define the decision tree and therefore the output. What is challenging in this scenario is producing quality samples from the posterior distribution on parameters. As the decision tree is nonlinear, evaluation of the posterior predictive distribution in Equation \ref{eq:bdt-pred} is analytically intractable, requiring approximation methods such as Monte Carlo integration. This in turn requires samples from the posterior distribution on parameters, as shown in Equation \ref{eq:predictive-dist-trees}. Therefore, producing useful samples from the posterior parameter distribution for the decision tree model is the main focus of this paper.

A major difficulty in producing samples from the decision tree posterior is that the model is transdimensional, meaning that the dimension of the parameter space changes. This issue is directly related to the topology parameter $\mathcal{T}$, in which the number of internal nodes $n$ and leaf nodes $n_\ell$ can vary, thus changing the size of the remaining parameter vectors. 
In this paper, we define and investigate using a novel strategy, entitled DCC-Tree, to handle this issue. This strategy was inspired by the work of \citet{zhou2020divide} within a probabilistic programming context and incorporates the benefits of HMC, as shown in \citet{cochrane2023rjhmctree}, to gain the advantage in sampling efficiency while simultaneously removing the observed computational overhead. The next section will detail the DCC-Tree sampling algorithm and the model likelihood and prior definitions used to apply Bayesian inference to the decision tree model.

\section{DCC-Tree Sampling Algorithm}
The DCC-Tree algorithm is based on the idea that the overall parameter space, which is defined by a joint discrete-continuous distribution, can be broken up into subspaces depending on the value of the discrete variable. The method progresses by considering a subset of all parameter values at each iteration. This section will first discuss the overall DCC-Tree approach to provide a contextual understanding of the algorithm. See Appendix \ref{sec:app-valid} for details on the correctness of the overall inference strategy.

\subsection{Algorithm Overview}\label{sec:dcc-background}

The premise behind the DCC-Tree algorithm is to consider the sample space as a union of disjoint subspaces of varying dimensions. The method exploits this underlying structure to the problem in an attempt to explore the posterior distribution of decision tree parameters. The method progresses as follows:
\begin{enumerate}
    \item The target distribution, which represents the parameter posterior distribution for the decision tree, is split up based on the different tree topologies.
    \item Local inference is run on each tree topology to generate a number of samples based on the corresponding decision tree parameters. 
    \item Each set of local samples is recombined based on the marginal likelihood of that tree topology to give an estimate of the overall space. 
\end{enumerate}
In this paper, we use the variable $m$ to refer to a distinct tree topology that defines a specific subspace, i.e. $m \equiv \mathcal{T}_m$, with the remaining local parameters collected into the vector $\Theta_m$. This implies that the tree topology parameter is discrete (and can be considered categorical, i.e. no natural numerical/ordering), whereas the remaining parameters are continuous. Note that trees with different topologies but the same number of leaf nodes $n_\ell$ are considered different subspaces under this definition. 

The decision tree posterior distribution can be divided into disjoint subspaces, which are easier to consider separately, then combined in such a way that the overall distribution is recovered. In order to coherently describe the overall parameter space, it is necessary for the DCC-Tree method that the parameter vector is all-encompassing, that is,
\begin{equation}
    \Theta = \left[ \begin{matrix}
        \Theta_1 &  \Theta_2 & \dots &  \Theta_M
    \end{matrix} \right].
\end{equation}
 Here, $ \Theta_m$ denotes the set of model parameters corresponding to the specific model structure $m$ which is indexed by the discrete random variable $\mathcal{M}$ with support $\{ 1, \dots, M \}$. Note that this definition is in contrast to the parameter vector with varying dimensions commonly assumed in other Bayesian decision tree methods.
 
 The target distribution is the posterior distribution on both the model parameters $\Theta$ and the discrete random variable $\mathcal{M}$. However, it is challenging to sample from this mixed continuous-discrete distribution as it jointly considers all model structures and parameters. It is much easier to draw samples from the distribution when considering each individual subspace. The DCC-Tree algorithm exploits this fact by generating samples from each subspace and combining these in an appropriate manner to recover samples from the overall joint distribution. 

If the discrete random variable takes on the value $m$, then the posterior distribution is proportional to the local parameter distribution  (see Appendix \ref{sec:app-valid}),
\begin{equation} \label{eq:dcc-joint}
    p(\Theta,m\mid \mathcal{D}) \propto p(\Theta_m\mid m,\mathcal{D}). 
\end{equation}
As a result, each tree topology can be considered separately with respect to the local inference method, and later combined appropriately to give an estimate of the overall space. This relationship can be used to provide an estimate of the distribution of the entire posterior parameter space, and as such, evaluate expectation integrals such as the posterior predictive distribution. The sign of proportionality in Equation \ref{eq:dcc-joint} is related to the marginal likelihood of the local parameter space. Therefore, both the local distribution $p(\Theta_m \mid \mathcal{D})$ and corresponding marginal likelihood $Z_m$ for each subspace $m$ need to be evaluated. Note that the nonlinearity of the decision tree model means that these terms will need to be approximated. 

HMC is used to generate samples to approximate the local distribution for each tree subspace. To achieve this, the soft-decision function and input selection definition (HMC-DFI) originally defined in \citet{cochrane2023rjhmctree} is adopted in this paper to take full advantage of the HMC sampling routine (details presented in Section \ref{sec:model-def}). This implies that each local parameter vector is described as 
\begin{equation}
    \Theta_m = (\bm{\Delta}_m,\bm{\tau}_m,\bm{\theta}_m).
\end{equation}
The marginal likelihood of each subspace $Z_m$ can be computed using a pseudo-importance sampling technique from these samples (further details on this will be provided in Section \ref{sec:marg-llh-calc}). Therefore, HMC provides the local inference method by which to approximate both the local parameter distribution $p(\Theta_m \mid \mathcal{D})$ and corresponding marginal likelihood $Z_m$ as required for the algorithm. 

The following sections will present the relevant information for the DCC-Tree algorithm, namely, the softened model definition which allows HMC to be applied, the marginal likelihood calculation, the local density estimation and the exploration of the tree structures. The overall algorithm will then be presented.

\subsection{Soft Decision Tree Model Definition}\label{sec:model-def}
As noted in \citet{cochrane2023rjhmctree}, one of the major difficulties in applying HMC to the decision tree model is the hard split parameterisation. To take full advantage of the HMC sampling method, we adopt the parameterisation used for the HMC-DFI method \cite{cochrane2023rjhmctree}. This section will summarise this information for the notation used in this paper.

Under the soft decision function and input selection specification, the splitting index is parameterised as a unit simplex and denoted $\Delta_{\eta_j}$ for internal node $\eta_j$, where
\begin{equation}
    \Delta_{\eta_j} = \left[ w_{\eta_j,1}, w_{\eta_j,2}, \dots, w_{\eta_j,n_x} \right], \quad w_{\eta_j,i} > 0,\quad \sum_{i=1}^{n_x}w_{\eta_j,i} = 1.
\end{equation}

Assume that the discrete random variable takes on the value $m$ with corresponding local parameter vector $\Theta_m  =(\bm{\Delta}_m,\bm{\tau}_m,\bm{\theta}_m).$. For a specific tree subspace $m$, the probability that a datapoint $(\mathbf{x}_i, y_i )$ goes to the left at internal node $\eta_j$ is defined as,
\begin{equation}\label{eq:psi-index}
    \psi(\mathbf{x}_i\mid \Theta_m,\eta_j) = f\left(\frac{\mathbf{x}_i\Delta_{m,\eta_j}-\tau_{m,\eta_j}}{h} \right),
\end{equation}
where $\Delta_{m,\eta_j}$ and $\tau_{m,\eta_j}$ denote the splitting index and splitting threshold respectively of the internal node $\eta_j$. Note that the logistic function $f(x) = \left( 1+\exp(-x)\right) ^{-1}$ and the method for varying the split parameter $h$ as discussed in \citet{cochrane2023rjhmctree} has again been adopted here. The probability that a datapoint is assigned to a specific leaf node can then be computed as the total probability along the path required to reach the leaf node. If we denote the probability of a datapoint $(\mathbf{x}_i, y_i )$ being assigned to leaf node $\eta_{\ell_k}$ in decision tree $m$ as $\phi_{i,m,k}$, the probability can be expressed as follows,
\begin{equation} \label{eq:phi}
    \phi_{i,m,k} (\mathbf{x}_i\mid \Theta_m,\eta_{\ell_k}) = \prod_{\eta \in \mathcal{A}(\eta_{\ell_k})}  \psi(\mathbf{x}_i\mid \Theta_m,\eta)^{R_{\eta}} (1-\psi(\mathbf{x}_i\mid \Theta_m,\eta))^{1-R_{\eta}},
\end{equation}
where $\mathcal{A}(\eta)$ and $ R_\eta$ denote the set of ancestor nodes and direction vector for node $\eta$.

\subsubsection*{Prior Specification}\label{sec:prior}

The prior on the joint mixed discrete-continuous distribution can be split into a prior on the discrete variable $\mathcal{M}$ and a prior on the continuous parameters $\Theta_m$ for each subspace.  The prior on the discrete variable is taken to align with the standard tree structure prior as originally defined in \citet{chipman1998bayesian}. This relates the probability of each tree subspace $m$ to the corresponding tree structure $\mathcal{T}_m$. Let $\I_m$ and $\Lf_m$ denote the set of internal and leaf nodes corresponding to the specific tree subspace $m$. The probability that the discrete random variable takes on the value $m$ is then defined as,
\begin{equation}\label{eq:cgm-prior} 
    p(m) \propto \prod_{\eta_j \in \I_m} p_{\textsc{split}}(\eta_j) \times \prod_{\eta_{\ell_k} \in \Lf_m} (1-p_{\textsc{split}}(\eta_{\ell_k})),
\end{equation}
where $p_{\textsc{split}}(\eta)$ denotes the probability that a node $\eta$ will split. Again, the definition used in \citet{chipman1998bayesian} is adopted here, where $p_{\textsc{split}} = \alpha(1+d_\eta)^{-\beta}$, with hyperparameters $\alpha \in (0,1)$ and $\beta \geq 0$ and where $d_\eta$ represents the depth of the node in the tree. 

The priors on the local parameters for each tree subspace are defined independently based on the specific decision tree model under consideration. If the discrete random variable relating to the tree structure takes on the value $m$, then the prior definition can be expressed as,
\begin{equation} \label{eq:tree-prior}
        p(\Theta_m) = p_\Delta(\bm{\Delta}_m\mid m) p_\tau(\bm{\tau}_m\mid m)   p_\theta(\bm{\theta}_m\mid m).
\end{equation}
The discrete variable $m$ refers to a specific tree topology $\mathcal{T}_m$ which dictates the dimension of the remaining parameters (i.e. $\bm{\Delta}_m/\bm{\tau}_m/\bm{\theta}_m$) through the number of internal nodes $n$ and leaf nodes $n_\ell$. Similarly to \citet{cochrane2023rjhmctree}, the priors on these continuous variables are defined to ensure the flexibility of the model and are taken to be,
\begin{align*}
    \Delta_{m,j} &\sim \text{Dir}(\bm{\alpha}), && \tau_{m,j} \sim \mathcal{B}(1,1), &&&& \eta_j \in \I_m, \quad j=1,\dots,n \\
    \mu_{m,k} &\sim \mathcal{N}(\alpha_\mu,\beta_\mu), && \sigma_m  \sim \Gamma^{-1}(\alpha_\sigma,\beta_\sigma), &&&& \eta_{\ell_k} \in \Lf_m,\quad k=1,\dots,n_\ell
\end{align*}
where Dir denotes the Dirichlet distribution, $\mathcal{B}$ the beta distribution, $\mathcal{N}$ the normal distribution and $\Gamma^{-1}$ the inverse-gamma distribution. 

\subsubsection*{Likelihood Definitions}

The likelihood for soft classification trees  is taken to be the Dirichlet-Multinomial joint compound as defined in Equation \ref{eq:llh-class},
\begin{equation}\label{eq:llh-class}
    \ell(\mathbf{Y}\mid \mathbf{X},\Theta_m,m) = \prod_{k=1}^{n_\ell} \left[ \frac{\Gamma(A)}{\Gamma(\Phi_k+A)} \prod_{c=1}^C\frac{\Gamma(\bm{\phi}_{c,k} +\alpha_m)}{\Gamma(\alpha_m)} \right],
\end{equation}
where $C$ refers to the number of output classes, $n_\ell$ is the number of leaf nodes in the tree structure $m$ and $\Phi_k = \sum_c \bm{\phi}_{c,k}$ for each leaf $\eta_{\ell_k}$. The variable $\bm{\phi}_{c,k} $ represents the probability of each datapoint $(\mathbf{x}_i,y_i)$, for which the output class is $y_i = c$, being assigned to leaf node $\eta_{\ell_k}$ in tree $\T$ and is expressed as
\begin{equation}\label{eq:temp}
    \bm{\phi}_{c,k}  = \sum_{i=1}^N\phi_{i,m,k}(\mathbf{x}_i\mid \T,\eta_{\ell_k})\mathbb{I}(y_i = c)
\end{equation}
where $\phi_{i,m,k}$ is as previously defined in Equation~\ref{eq:phi}.

Following \citet{linero2018bayesian}, the likelihood for regression trees in the soft setting is defined to be, 
\begin{equation}\label{eq:llh-reg}
    \ell\left(\mathbf{Y} \mid \mathbf{X},\Theta_m,m\right) = \prod_{i=1}^N\left(2\pi\sigma^2\right)^{-\frac{1}{2}}\exp\left[ -\frac{1}{2\sigma^2} \left(  \sum_{k=1}^{n_\ell} \phi_{i,m,k}(\mathbf{x}_i)\cdot(\mu_{m,k} - y_i) \right) ^2 \right],
\end{equation}
where again $\phi_{i,m,k}$ is as defined in Equation~\ref{eq:phi}, $\mu_{m,k}$ is the mean value of the assumed normal distribution within each leaf node $\eta_{\ell_k} \in \Lf_m$ and $\sigma_m$ is the assumed constant variance across all leaf nodes for tree subspace $m$. 

\subsection{Exploring Tree Structures}

A challenging aspect of applying Bayesian inference to decision trees is adequately exploring the different tree structures. It is desirable to spend most of the computational effort considering trees with high posterior probability in order to better approximate the overall density. The approach taken in this algorithm uses a global random-walk scheme to propose new trees and a utility metric to select the next tree to be considered. 

A tree-specific set of proposals is used to propose new tree topologies, assisted by the fact that each tree is uniquely defined based on which nodes are in the set of leaf nodes. To ensure the correctness of the method, the standard grow/prune/stay random walk method is used to propose tree topologies such that there is a non-zero probability of exploring any topology. This global proposal scheme is identical to that used for the transdimensional transitions in \citet{cochrane2023rjhmctree}.

The standard method in literature by which to propose a tree structure to be considered is to either stay or randomly select from a valid set of nodes to either grow or prune. However, this method does not account for the relative likelihood of each tree within the proposal and may miss some trees if the posterior is multimodal. Instead, a utility function is adopted in this algorithm as a means of proposing the next tree structure to consider as a way to improve the exploration of the posterior distribution. This is based on multiple important considerations including how many times the tree has already been considered and the marginal likelihood of the tree. 

When a tree topology is first considered, the No-U-Turn-Sampler (NUTS) algorithm \cite{hoffman2014no} is run to initialise the sampling method for that topology. This is an extension of the standard HMC sampling algorithm which adapts the hyperparameters of the method. Note that this is in contrast to the greedy up-hill MCMC initialisation used in \citet{zhou2020divide}. After the burn-in phase, a set of samples is collected and used to compute the marginal likelihood estimate $\hat{Z}_m$. This in turn is used in the utility function calculation, which is then used to select the next tree for local inference. 

The utility function that has been implemented originates from \citet{rainforth2018inference} and is defined for tree subspace $m$ as,
\begin{equation}\label{eq:utility}
    U_m := \frac{1}{S_m}\left( \frac{(1-\delta)\hat{\tau}_m}{\max_m\{\hat{\tau}_m\} } + \frac{\delta\hat{\rho}_m}{\max_m\{\hat{\rho}_m\}} + \frac{\beta \log \sum_m S_m }{\sqrt{S_m}} \right)
\end{equation}
where $S_m$ is the number of times local inference has been performed on tree subspace $m$; $\hat{\tau}_m$ denotes the exploitation term that places a higher value on trees that have a high marginal likelihood, or that have high variance in the samples (and thus may need to be explored further); $\hat{\rho}_m$ is the exploration term used to estimate whether there may be important future samples at a specific tree subspace.  Calculations of these terms are described further in Appendix \ref{sec:app-exploit}. There are also two user-specified parameters: $0 \leq \delta \leq 1$ is a hyperparameter controlling the trade-off between exploration and exploitation; $\beta > 0$ is the standard optimism boost hyper-parameter \cite{zhou2020divide}.

\subsection{Marginal Likelihood Calculation}\label{sec:marg-llh-calc}

In the DCC-Tree algorithm, samples are generated within each individual tree subspace via the NUTS algorithm to provide an estimate of the local distribution. However, each subspace may not be equally important to the overall parameter space and must be assigned an associated weight. This weighting is related to the marginal likelihood of the local distribution. Some approximate inference methods, such as SMC, provide relative weights for each sample and therefore can easily produce an estimate of the marginal likelihood. However, MCMC-based methods such as HMC use sequential samples to estimate the posterior distribution via an accept-reject technique \cite{metropolis1953equation, hastings1970monte}. As such, each sample is assumed to contribute equally to the estimate and therefore doesn't contain any relative weight information. The method by which the marginal likelihood is estimated using MCMC-based samples is discussed in this section.

\subsubsection*{Layered Adaptive Importance Sampling}
\vspace{-0.5em}
The marginal likelihood for each tree topology is estimated via the IS-after-MCMC method, layered adaptive importance sampling (see Section 5.4 of \citet{llorente2023marginal} for further details). Intuitively, this method attempts to create pseudo-samples to be used as importance samples from which the marginal likelihood can be approximated. 

Consider a specific tree subspace $m$. Let the set of samples generated via HMC for this subspace be denoted $\nu_m^{(i,j)} = \{\bm{\Delta}_m^{(i,j)},\bm{\tau}_m^{(i,j)},\bm{\theta}_m^{(i,j)}\}$ for $i = 1,\dots,N_T$ total number samples for each $j=1,\dots,N_c$ parallel chains. The estimate uses each sample $\nu_m^{(i,j)}$ to define a proposal distribution $q_{i,j,m}(\xi \mid \nu_m^{(i,j)},\Sigma_j)$ from which $k=1,\dots,N_M$ pseudo-importance samples $\xi^{(i,j,k)}$ are drawn. Here, $\nu_m^{(i,j)}$ acts as the mean value and $\Sigma_j$ as a covariance matrix. These pseudo-importance samples are then used to produce an estimate of the marginal likelihood via
\begin{equation}\label{eq:marg-llh-calc}
    \hat{Z}_m = \frac{1}{N_TN_cN_M} \sum_{i=1}^{N_T} \sum_{j=1}^{N_c} \sum_{k=1}^{N_M} \tilde{w}_m^{(i,j,k)},  \quad \tilde{w}_m^{(i,j,k)} = \frac{\tilde{p}_m(\xi^{(i,j,k)})}{\Phi_m(\xi^{(i,j,k)})},
\end{equation}
where $ \tilde{p}_m(\xi^{(i,j,k)} )$ is the unnormalised posterior distribution on parameters (note this also includes priors on parameters and tree structure) for tree subspace $m$, evaluated at each pseudo-importance sample $\xi^{(i,j,k)}$. The term $\Phi_m(\xi^{(i,j,k)})$ is also computed using these pseudo-importance samples (note that this can change depending on which definition out of Equations (142) - (145) is selected from Table 5.5 of \citet{llorente2023marginal}), and is taken here to be either the basic definition,
\begin{equation}
    \Phi_m(\xi^{(i,j,k)}) = q_{i,j,m}(\xi^{(i,j,k)} \mid \nu_m^{(i,j)},\Sigma_j).
\end{equation}
or the spatial definition,
\begin{equation}
    \Phi_m(\xi^{(i,j,k)}) = \frac{1}{N_T} \sum_{n=1}^{N_T} q_{n,j,m}(\xi^{(i,j,k)} \mid \nu_m^{(n,j)},\Sigma_j),
\end{equation}
where the definition of the proposal distributions $q$ will be discussed in the next section. Note that in practice, the log marginal likelihood is estimated as opposed to the marginal likelihood itself (see Appendix \ref{sec:app-log-marg} for details).

\subsubsection*{Proposal Distributions} 
The proposal density is defined jointly for all parameters within a given tree topology. Due to the possible multi-modality of the posterior distribution (more discussion on this in Section \ref{sec:subspaces}), the covariance of each proposal density is calculated only on the within-chain samples (not cross-coupled with other parallel chains). 
Further, the covariance matrix is calculated using the within-chain samples for that iteration of local inference. 

Note, however, that the splitting thresholds $\bm{\tau}_m$, splitting simplexes $\bm{\Delta}_m$ and leaf node variance $\sigma_m$ (if considering regression problems) for each subspace are all constrained variables. The proposal distribution must therefore be defined in the unconstrained space, otherwise, this could lead to proposals that are not within the support of the original distribution. Conversion from the constrained space to the unconstrained space requires particular consideration of the change of variables, which impacts the proposal distribution computation.

Let $f$ be a differentiable and invertible function that defines the transformation from the constrained parameters space to the unconstrained space. Let $X$ and $Y$ be random variables in the constrained and unconstrained space respectively. Under the transformation $Y = f(X)$, the multivariate change of variable theorem means that the distribution of the variables in the unconstrained space is defined to be,
\begin{equation}
    p_Y(y) = p_X\left(f^{-1}(y)\right) \left| \det J_{f^{-1}}(y) \right|
\end{equation}
where $p_{Y}$ and $p_{X}$ denote the distribution in the unconstrained and constrained space respectively \cite{stan2019}. That is, the distribution in the unconstrained space is scaled by the absolute Jacobian determinant of the transformation. 

The pseudo-importance samples for each parameter are drawn from a multivariate normal distribution centered around the original samples $\nu_m^{(i,j)}$ transformed to the unconstrained space,
\begin{equation}
    \xi^{(i,j,k)} \sim \text{MVN}\left(f\left(\nu_m^{(i,j)}\right),\Sigma_j\right).
\end{equation}
Here, MVN denotes the multivariate-normal distribution, $\Sigma_j$ is the parameter covariance of chain $j$ and $f$ is the transformation to the unconstrainted space. Note that the dimension of the multivariate normal and covariance matrix depends on the selected tree $m$, which has been dropped from the above notation. The term $\Phi_m$ is evaluated by accounting for this change of variable transformation (either the logit, stick-breaking, or log transformation for $\bm{\tau}_m$, $\bm{\Delta}_m$ or $\sigma_m$ respectively) when evaluating the proposal distribution $q$ as each pseudo-importance sample. 

\subsection{Local Density Estimation}\label{sec:subspaces}
The NUTS algorithm uses the burn-in phase to adapt relevant hyperparameters to the sampling algorithm, while simultaneously encouraging movement to areas of high likelihood. However, it is possible that after this burn-in phase, samples corresponding to different chains within the same tree subspace may explore different modes. This is due to the different initialisation values used for each chain during the burn-in phase and the difficult geometry of the decision tree posterior. As a result, there is a potential multi-modality of each local parameter distribution $p_m(\Theta_m)$.

Furthermore, these modes may have different posterior masses within the subspace. This motivates using the pseudo-importance samples with normalised weights to provide an estimate of the local posterior subspace. This will have the beneficial effect of reducing the impact of any chains with lower posterior density on the final overall density estimate. Equation \ref{eq:post-est} shows how the weighted importance samples can be used to approximate the posterior for tree subspace $m$,
\begin{equation}\label{eq:post-est}
    \hat{p}_m(\Theta_m ) = \sum_{i=1}^{N_T} \sum_{j=1}^{N_c}  \sum_{k=1}^{N_M} w_m^{(i,j,k)} \delta_{\xi^{(i,j,k)}}(\cdot) , \quad w_m^{(i,j,k)} = \frac{\tilde{w}_m^{(i,j,k)}}{\sum_{i=1}^{N_T} \sum_{j=1}^{N_c}\sum_{k=1}^{N_M} \tilde{w}_m^{(i,j,k)}}
\end{equation}
where $\delta(\cdot)$ is the Dirac delta function and the unnormalised weights $\tilde{w}_m^{(i,j,k)}$ are as defined in Equation \ref{eq:marg-llh-calc}.

\subsection{Overall Algorithm}

The overall DCC-Tree sampling algorithm, shown in Algorithm \ref{alg:dcc-tree}, follows a similar implementation to that described in \citet{zhou2020divide}. The method starts by generating an initial set of tree topologies $\mathcal{T}_m$ from the prior distribution, keeping track of those proposed and the corresponding number of times proposed $C_m$. Any tree topologies that have been selected more than a user-specified threshold $C_0$ are then stored in the set of currently active trees. Note that a maximum number of trees to keep track of, $T_\textsc{max}$, can also be specified, which will remove the tree with the lowest utility value (as defined by Equation \ref{eq:utility}).

Once a new tree is added to the set of active trees, the NUTS method is run to initialise each of the $N_c$ independent parallel chains using $N_{\textsc{init}}$ burn-in samples. This initialises the sampling method by adapting hyperparameters, such as the mass matrix and step size, and also provides a good starting point for subsequent local inference. The active tree with the highest utility value is then selected for local inference, in which an additional $N_s$ HMC samples are generated for each of the $N_c$ parallel chains. Using the total number of local inference samples $N_T$ (which includes the new $N_s$ samples) the marginal likelihood of the current tree can be computed via Equation \ref{eq:marg-llh-calc}.

After local inference is complete, a global updating step is performed in which a new tree topology is proposed -- either stay, grow, or prune -- based on the current topology. If tree topology is already in the set of discovered trees, then the number of times proposed $C_m$ is incremented, otherwise, it is added to the set. The algorithm continues to run for $T$ global iterations after which a set of samples that approximates the local posterior distribution and an estimate of the log marginal likelihood for each explored subspace is returned. The overall algorithm is shown in Algorithm \ref{alg:dcc-tree}. 
\begin{algorithm}
    \DontPrintSemicolon
    \SetAlgoLined
    \SetKwComment{tcb}{(}{)}
    \SetKwInput{KwInit}{Init}
    \KwIn{Number of iterations $T$, number of initial trees to draw from prior $T_0$, number of parallel chains $N_c$, number of local samples per iteration $N_s$, number of times tree is proposed before being added to active trees $C_0$, number of burn-in samples for each subspace $N_{\textsc{init}}$. }
    \KwInit{Set of discovered trees $\mathbb{D} = \emptyset$, set of currently active trees $\mathbb{A} = \emptyset$.}
    Generate $T_0$ trees from the prior and store $\mathbb{D} = \{\T_m\}$. \;
    \Indp\For{$t = 0 $ \KwTo $T$}{
        \uIf{any $\T_m \in \mathbb{D}$ selected more than $C_0$ times}{
        Add $\T_m$ to set of active trees $\mathbb{A}$.\;}
        \uIf{any new $\mathbb{T}_m \in \mathbb{A}$}{
        Run $N_{\textsc{init}}$ burn-in samples for each of the $N_c$ parallel chains via standard HMC adaption phase procedure (increasing $N_{init}$ as necessary). \;}
    1. Calculate utility $U_m$ for each $\T_m \in \mathbb{A}$ via Equation \ref{eq:utility} -- select $\mathbb{T}_m$ with highest $U_m$ to perform local inference. \;
    2. Generate $N_s$ additional samples for each $N_c$ parallel chain via HMC for the selected tree $\mathbb{T}_m$. \;
    3. Calculate marginal likelihood estimate $\hat{Z}_m$ of tree $\mathbb{T}_m$ via Equation \ref{eq:marg-llh-calc}.\;
    4. Apply global update via grow/prune/stay method to selected $\mathbb{T}_m$ to propose  $\mathbb{T}^*$. \;
    \uIf{$\mathbb{T}^* \in \mathbb{D}$}{
    Increase by one the number of times selected $C_{m^*}$ for $\mathbb{T}^*$. \;
    } 
    \Else{
    Add $\mathbb{T}^*$ to  $\mathbb{D}$. \;}.
    }
    \Indm
    \Indp\For{$m = 0 $ \KwTo $M=$len$(\mathbb{A})$}{
        Compute estimate of $\hat{p}_m(\Theta_m)$ via Equation \ref{eq:post-est}. \;
    }
    \Indm
    \KwOut{Generated samples from each decision tree subspace and the corresponding marginal-likelihood estimate $\{ \hat{p}_m(\Theta_m),\hat{Z}_m\}_{m=1}^M$.}
    \caption{DCC-Tree Sampling Algorithm}
    \label{alg:dcc-tree}
\end{algorithm}

\section{Experiments}\label{sec:dcc-results}

The DCC-Tree algorithm was tested on a range of synthetic and real-world datasets commonly used in the Bayesian decision tree and machine learning literature. Each method was compared based on either the mean-square error or the accuracy across the testing and training datasets for regression and classification problems respectively.  

In a different manner to the MCMC-based methods, metrics for the DCC-Tree algorithm are computed after the entire algorithm is complete via the predictive posterior distribution. For a datapoint $(\mathbf{x}_j^*,y_j^*)$ in the testing dataset, the posterior predictive distribution is computed (with the appropriate simplifications due to independence) via the following expression,
\begin{equation}\label{eq:pred-dist-data}
    p\left(y_j^* \mid \mathbf{x}_j^*, \mathcal{D}\right) = \sum_{m=1}^M \int p\left(y_j^* \mid \mathbf{x}_j^*, \Theta, m \right) p\left(\Theta, m \mid\mathcal{D}\right) d \Theta.
\end{equation}
Monte Carlo integration is used to approximate this integral as follows,
\begin{equation}
    p(y_j^* \mid \mathbf{x}_j^*, \mathcal{D}) \approx \frac{1}{L} \sum_{i=1}^L p(y_j^* \mid \mathbf{x}_j^*, \Theta_{m^i}^i), \qquad (m^i, \Theta_{m^i}^i) \overset{\text{i.i.d.}}{\sim} \bar{w}_m\, p_m(\Theta_m).
\end{equation} 
Again, due to the soft split assumption, the output from each tree is a weighted sum of distributions. That is, given a tree, the distribution on the predicted output $y_j^*$ is given by
\begin{equation}
    p\left(y_j^* \mid \mathbf{x}_j^*, \Theta_{m^i}^{i}\right) = \sum_{k=1}^{n_\ell} \phi_{i,m,k} \cdot p\left(y_j^* \mid \mathbf{x}_j^*, \theta_{k,m}^{i}\right),
\end{equation}
where $\theta_{k,m}^{i}$ denotes the parameters associated with leaf node $k$ in sample $\Theta_{m^i}^{i}$. The distribution $p(y_j^* \mid \mathbf{x}_j^*, \theta_{k,m}^{i})$ is either a Gaussian distribution for regression or a categorical distribution for classification. 

For classification datasets, the accuracy is computed via
\begin{equation}
    \text{accuracy} = \frac{1}{N_\textsc{test}}\sum_{j=1}^{N_\textsc{test}} \mathbb{I}\left( y_j^* = \hat{y}_j^* \right), \quad \hat{y}_j^* = \text{argmax} \phantom{\cdot} p\left(y_j^* \mid \mathbf{x}_j^*,  \mathcal{D}\right),  
\end{equation}
and for regression problems, the mean-square-error is computed as,
\begin{equation}
    \text{MSE} = \frac{1}{N_\textsc{test}}\sum_{j=1}^{N_\textsc{test}} \left( y_j^* -  \hat{y}_j^* \right)^2, \quad \hat{y}_j^* = \mathbb{E}_{p(y_j^* \mid \mathbf{x}_j^*,\mathcal{D})}[y_j^*].
\end{equation}
Note that here the prediction is taken with respect to the entire posterior distribution $p\left(y_j^* \mid \mathbf{x}_j^*,\mathcal{D}\right)$.

\subsection{Bayesian Tree Synthetic Datasets}
The DCC-Tree algorithm was tested against two synthetic datasets common in the Bayesian decision tree literature, namely, the datasets from \citet{chipman1998bayesian} and \citet{wu2007bayesian}, which will be referred to as CGM and WU respectively. Details of these datasets are presented in Appendix \ref{sec:app-true-trees}. The performance was compared to other Bayesian decision tree methods using the same hyperparameters as in \citet{cochrane2023rjhmctree} to remain consistent, with those for the WU dataset provided in Table \ref{tab:hyperparams} in Appendix \ref{sec:app-hyp}. The DCC-Tree algorithm was run for $T=500$ iterations, with $N_s = 100$ local inference samples per iteration. Each new tree subspace was initialised using $N_{\textsc{init}} = 2000$ burn-in samples for the WU dataset and $N_{\textsc{init}}=5000$ for the CGM dataset. For each, $N_M=10$ pseudo-importance samples were drawn to compute the marginal likelihood estimate. The DCC-Tree method was run 10 times for different initialisation values with metrics averaged across all runs. A relatively non-informative prior on the tree structure ($\alpha_{\textsc{split}}=0.95$, $\beta_{\textsc{split}} = 1.0$), with the values of the split hyperparameters $h_{\textsc{init}}$ and $h_{\textsc{final}}$ summarised in Table \ref{tab:hyperparams-dcc} in Appendix \ref{sec:app-hyp}. 
The results for the DCC-Tree algorithm and comparison to other existing methods for the WU and CGM datasets are shown in Table \ref{tab:dcc-synth-metrics}.

\begin{table}[ht]
    \caption{Testing and training MSE for various methods when tested on the synthetic datasets of \citet{chipman1998bayesian} and \citet{wu2007bayesian}. }
    \label{tab:dcc-synth-metrics}
    \centering
    \begin{scshape}
    \begin{small}
        \resizebox{\textwidth}{!}{
            \begin{tabular}{lrcccccc}
            \toprule
             & & CGM  & WU & HMC-DF & HMC-DFI & DCC-Tree\\
            \midrule
            \multirow{2}{*}{CGM}    & Train MSE & 0.043(1.9e-4)  & \textbf{0.042(5.8e-5)}  & 0.043(2.0e-4)  &  0.043(1.9e-4)   & 0.043(5.2e-5) \\
                                    &  Test MSE & 0.064(0.014)  &  0.062(0.001) &  0.041(4.6e-4) & 0.041(4.5e-4)  &  \textbf{0.040(1.4e-4)}  \\
            \midrule
            \multirow{2}{*}{WU}     &  Train MSE & 0.059(2.3e-4) &  \textbf{0.054(1.4e-3)}  &   0.060(7.4e-4)  & 0.060(4.4e-4)   & 0.058(3.3e-5)  \\
                                    &  Test MSE & 0.112(0.034) &  0.073(0.038)   &   0.059(2.1e-3)   &  0.060(2.5e-3)    &  \textbf{0.059(5.3e-4)}   \\
            \bottomrule
            \end{tabular}
        }
    \end{small}
\end{scshape}
\end{table}

The DCC-Tree algorithm exhibits the best testing performance across the different methods for both datasets, although only just better than the other HMC-based methods. Notably, the variance of the DCC-Tree method is lower than other methods, and in some cases, by nearly an order of 10. Both CGM and WU methods show signs of overfitting on the two datasets, clearly noted when comparing the difference in training and testing performance. In fact, the WU method performs best on the training data for both datasets but is much worse than all HMC-based methods with respect to the testing performance. 

\begin{figure}
    \centering
    \begin{subfigure}[t]{0.8\textwidth}
      \includegraphics[width=\linewidth]{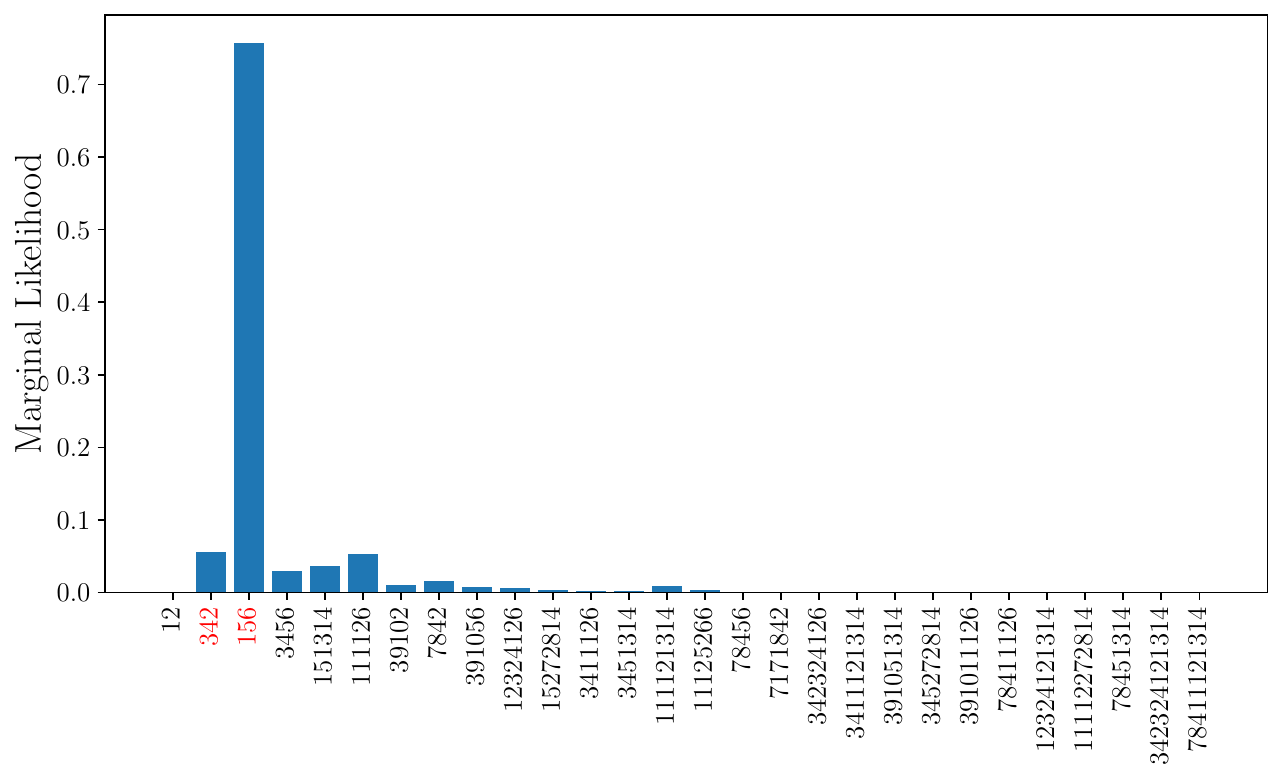}
      \caption{Marginal likelihood of tree structures for the WU dataset.}
    \end{subfigure}
    \\
    \medskip
    \begin{subfigure}[t]{0.8\textwidth}
      \includegraphics[width=\linewidth]{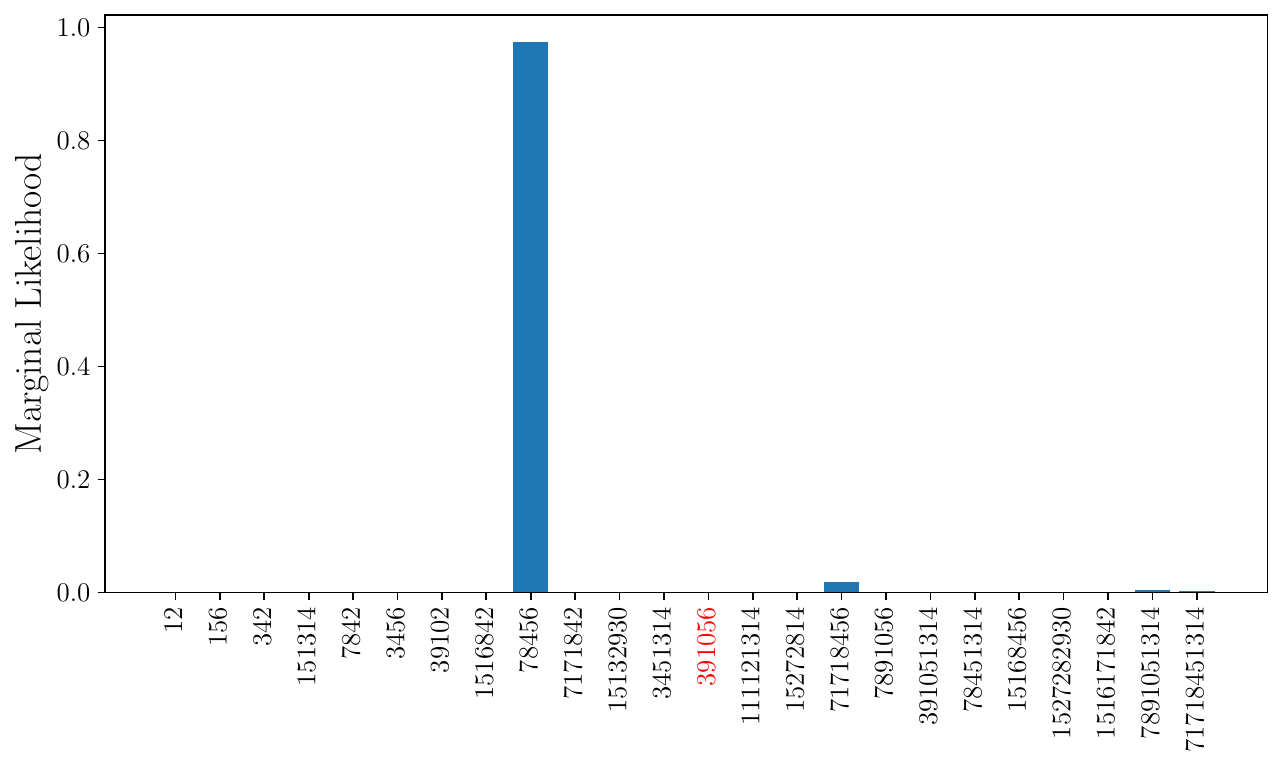}
      \caption{Marginal likelihood of tree structures for the CGM dataset.}
    \end{subfigure}
    \caption{Marginal likelihood of each tree structure for the DCC-Tree method. The tree structure used to generate the data is highlighted in red.}\label{fig:dcc-marg-llh-cgm-wu}
  \end{figure}

The marginal posterior distribution on the tree structure was also visualised for the DCC-Tree method, as defined by the marginal likelihood estimate $\hat{Z}_m$ for each tree subspace. Figure \ref{fig:dcc-marg-llh-cgm-wu} shows this for both the WU and CGM datasets for a single run, with the label for the true tree structure highlighted in red.   
Interestingly for the decision tree model, the likelihood of the data does not significantly decrease with additional nodes once the underlying true tree structure has been found. Instead, the penalty for larger than necessary tree structures comes from the prior. Using a relatively non-informative prior on the tree structure meant that the marginal likelihood estimate was affected mainly by the likelihood, and shows that the DCC-Tree method is somewhat insensitive to the prior definition. However, Figure \ref{fig:dcc-marg-llh-cgm-wu} does illustrate that this can be an issue when considering the marginal likelihood plot for the CGM dataset. The tree structure used to generate the data has a very low posterior probability. Instead, a tree structure with the same number of leaf nodes and which provides the same partition of the data has a posterior probability close to one. Further investigation into this issue showed that none of the parallel chains used for local inference discovered the correct mode for the true tree, resulting in a much lower marginal likelihood.

\begin{figure}[hbtp]
    \centering
    \begin{subfigure}[t]{.42\textwidth}
        \includegraphics[width=\linewidth]{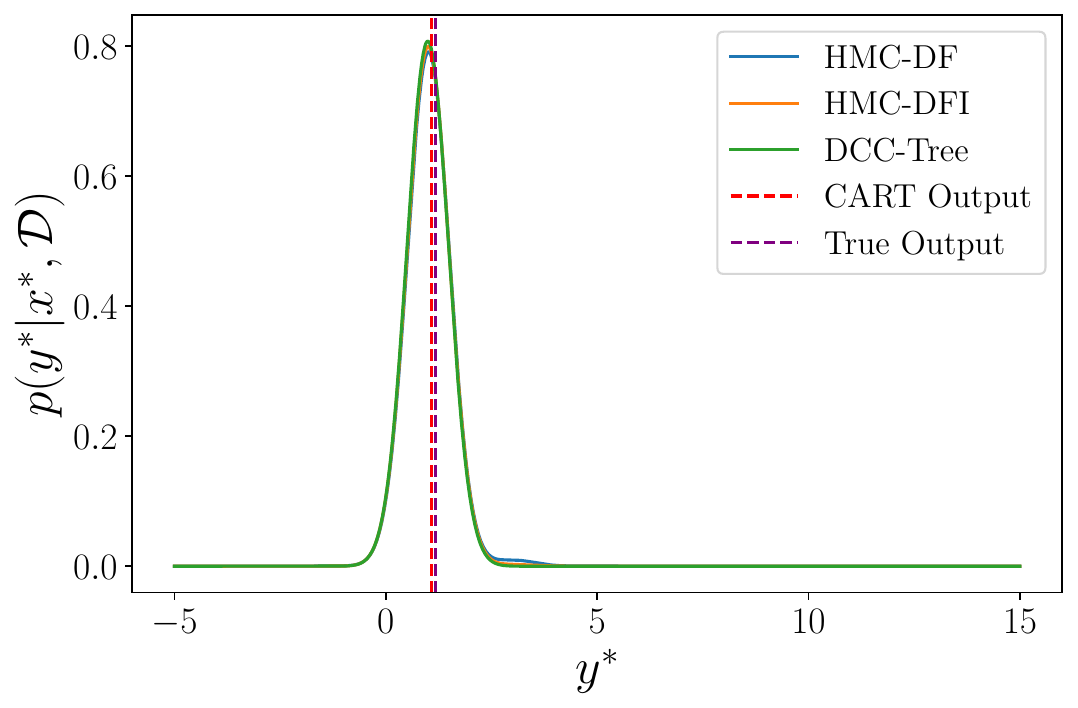}
      \end{subfigure}
      \begin{subfigure}[t]{.42\textwidth}
        \includegraphics[width=\linewidth]{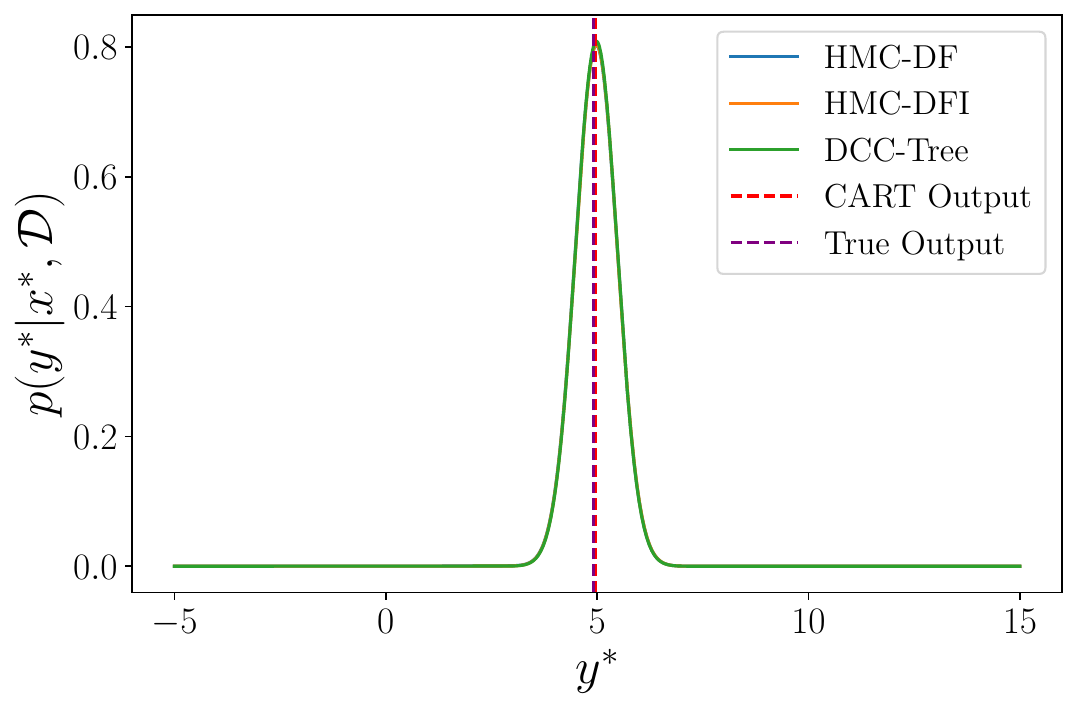}
      \end{subfigure}
      \begin{subfigure}[t]{.42\textwidth}
        \includegraphics[width=\linewidth]{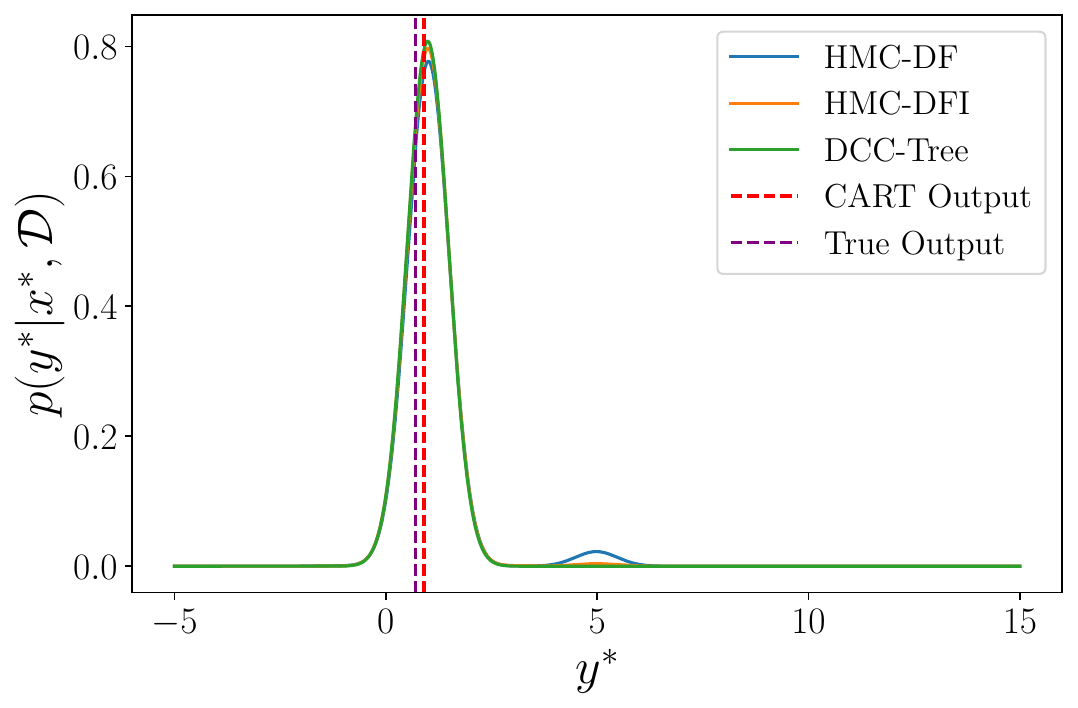}
      \end{subfigure}
      \begin{subfigure}[t]{.42\textwidth}
        \includegraphics[width=\linewidth]{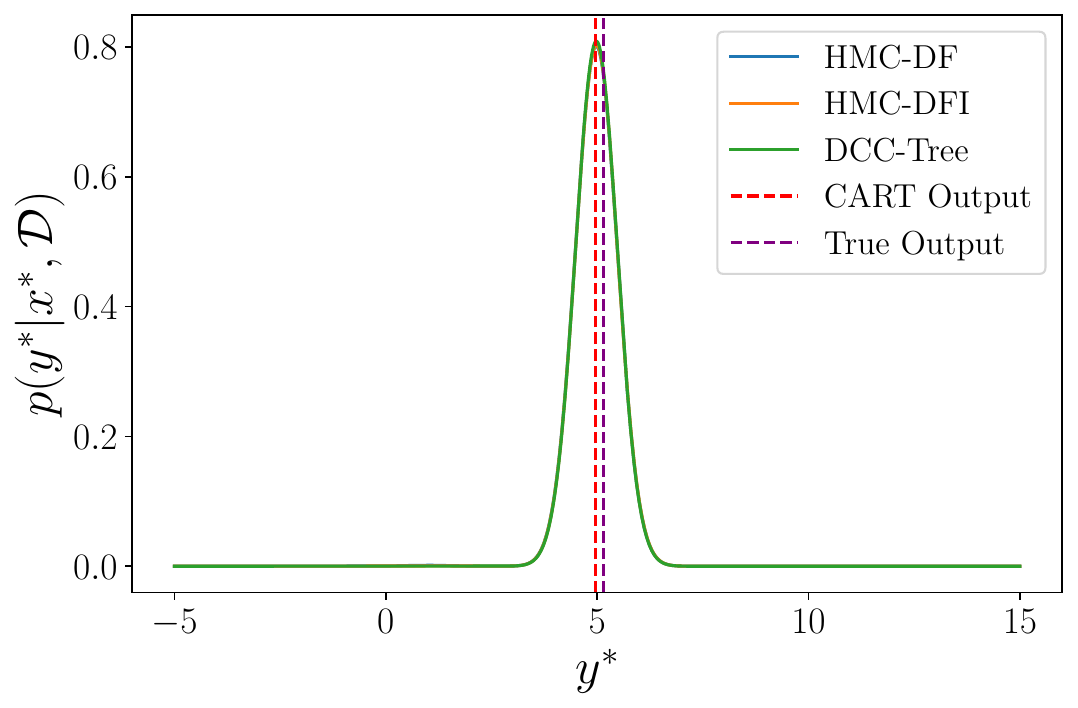}
      \end{subfigure}
      \caption{Posterior predictive distributions for four randomly selected testing datapoints from the synthetic WU dataset \cite{wu2007bayesian}.}\label{fig:dcc-wu-pred-dists}
  
    \vspace*{\floatsep}
  
    \begin{subfigure}[t]{.42\textwidth}
        \includegraphics[width=\linewidth]{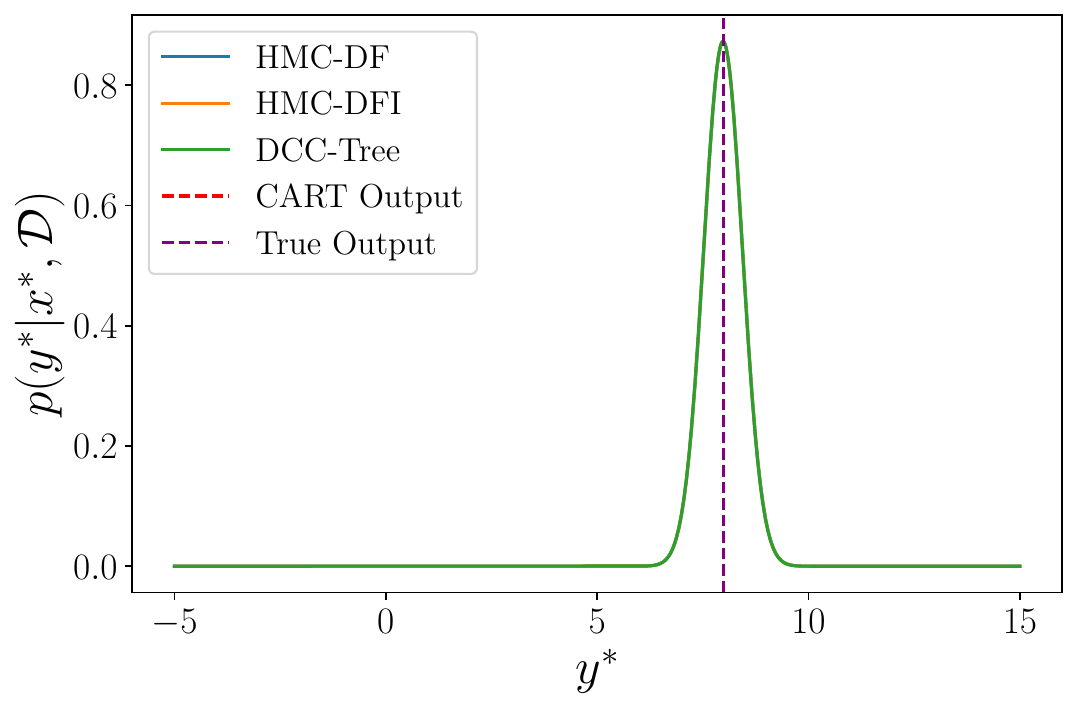}
      \end{subfigure}
      \begin{subfigure}[t]{.42\textwidth}
        \includegraphics[width=\linewidth]{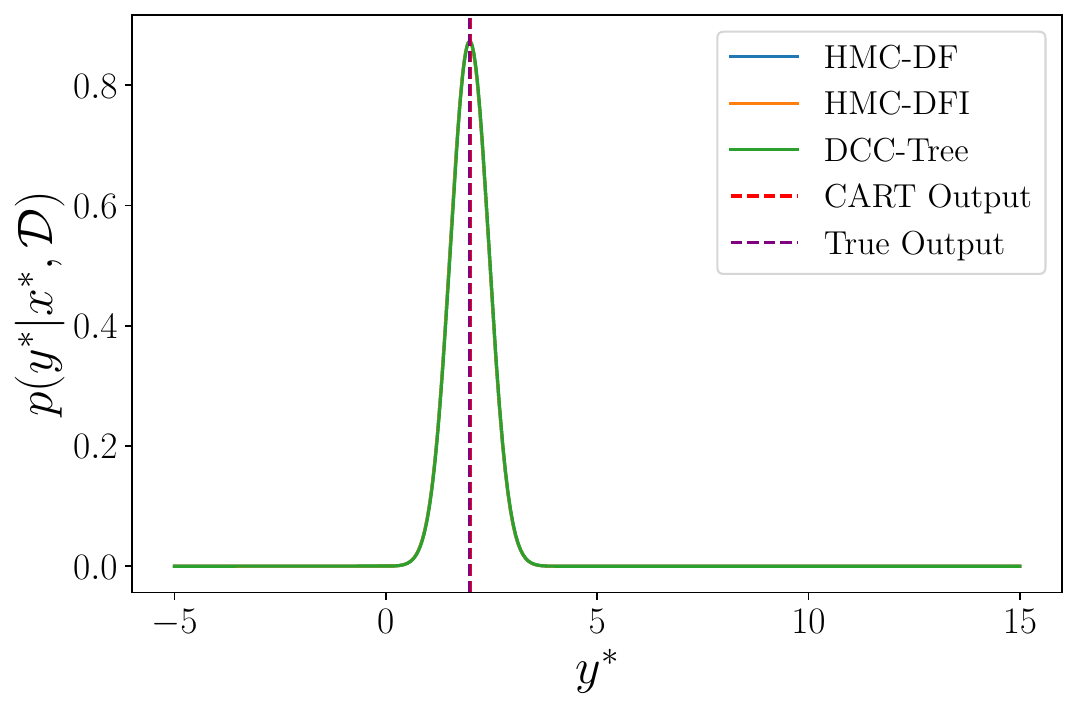}
      \end{subfigure}
      \begin{subfigure}[t]{.42\textwidth}
        \includegraphics[width=\linewidth]{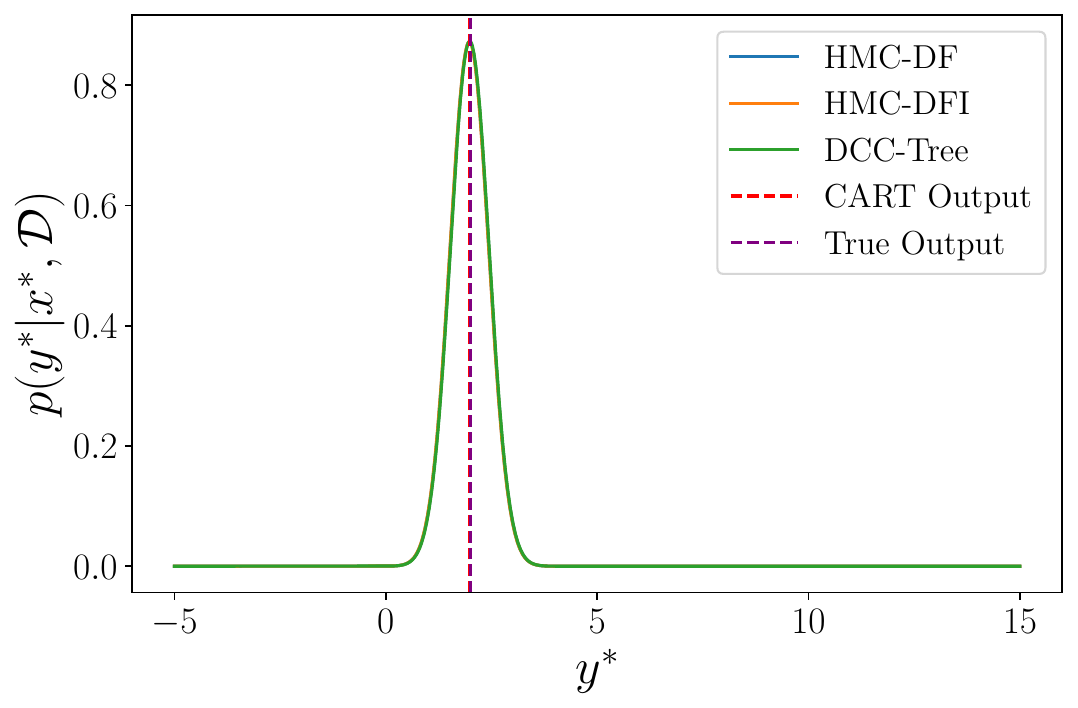}
      \end{subfigure}
      \begin{subfigure}[t]{.42\textwidth}
        \includegraphics[width=\linewidth]{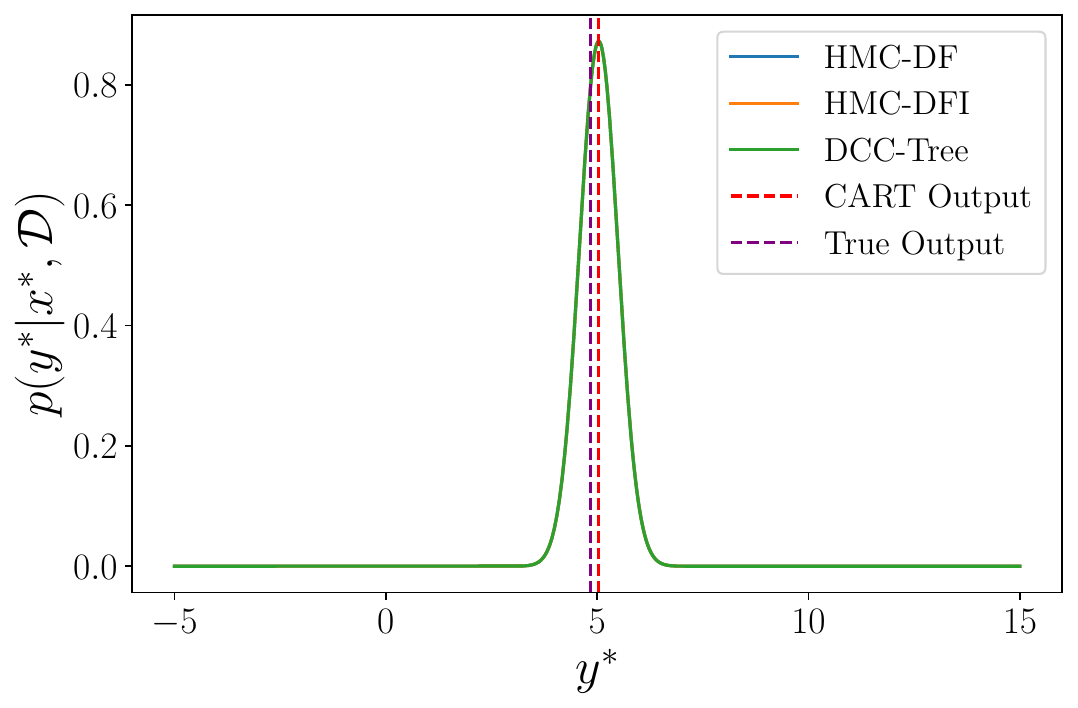}
      \end{subfigure}
      \caption{Posterior predictive distributions for four randomly selected testing datapoints from the synthetic CGM dataset \cite{chipman1998bayesian}.}\label{fig:dcc-cgm-pred-dists}
  \end{figure}

It is straightforward to plot the posterior predictive distribution for both the RJHMC-Tree and DCC-Tree methods as leaf parameters are not marginalised out and are therefore easily accessible for each method. For the CGM and WU datasets, four testing datapoints were randomly selected for which to compute the posterior predictive distributions for the RJHMC-Tree methods and the DCC-Tree method. Figure \ref{fig:dcc-wu-pred-dists} shows the posterior predictive distribution for four randomly selected testing datapoints in the WU dataset, and similarly in Figure \ref{fig:dcc-cgm-pred-dists} for the CGM dataset. The predictive distributions are compared to the predicted output from the standard CART model and the true output. 
It can be seen that in all cases, the distributions produced by the RJHMC-Tree methods and the DCC-Tree method line up nearly exactly. Furthermore, each strongly predicts the true value, in that the posterior distributions are highly peaked around the true output.

\subsection{Real-world Datasets}
The DCC-Tree method was also tested against a range of real-world datasets: Iris, Breast Cancer Wisconsin (Original), Wine, and Raisin Datasets (all available from \citet{Dua:2019}) and compared to other Bayesian decision tree methods. 

The DCC-Tree method was run for $T=500$ iterations with $N_s=100$ local inference samples generated per iteration. Split hyperparameters $h_{\textsc{init}}$ and $h_{\textsc{final}}$ used for each dataset are summarised in Table \ref{tab:hyperparams-dcc} in Appendix \ref{sec:app-hyp} with a relatively uninformative prior on the tree structure again used. Training and testing metrics were averaged across 10 runs of the DCC-Tree method with different initialisation values. The results are shown in Table \ref{tab:dcc-real-metrics}, along with the results for the other Bayesian tree methods. The standard deviation across the different runs is presented in parentheses. The best-performing method for each dataset is shown in bold.

\begin{table}[ht]
    \caption{Comparison of metrics for various methods for different real-world datasets.}
    \label{tab:dcc-real-metrics}
    \centering
    \begin{scshape}
    \begin{small}
        \resizebox{\textwidth}{!}{
    \begin{threeparttable}
        \begin{tabular}{lrcccccccc}
            \toprule
             & & CGM  & SMC & WU & HMC-DF & HMC-DFI & DCC-Tree\\
            \midrule
             \multirow{2}{*}{BCW}  &  Train Acc. & 0.983(0.004)  & \textbf{0.987(0.004)} & 0.978(0.005) & 0.973(0.005) & 0.981(0.003) & 0.982(0.001)    \\
             &  Test Acc.  &  0.939(0.014)  &  0.924(0.010) & 0.922(0.017)  &  0.940(0.010) & 0.952(0.007) & \textbf{0.952(0.004)} \\
            \midrule
            \multirow{2}{*}{Iris}    & Train Acc. &  \textbf{0.985(0.007)}  & 0.981(0.004)  & \multirow{2}{*}{N/A\tnote{1}}  & 0.977(0.010) & 0.975(0.009)  & 0.981(0.000)\tnote{2} \\
                &  Test Acc.  & 0.908(0.022)  &  0.909(0.022) &   & 0.906(0.026) & \textbf{0.917(0.023)} & 0.911(1.2e-16)\tnote{2} \\
            \midrule
            \multirow{2}{*}{Wine}    & Train Acc. & 0.957(0.016) & \textbf{0.985(0.011)}  &  \multirow{2}{*}{N/A\tnote{1}} & 0.949(0.021) & 0.952(0.016) & 0.958(0.010) \\
                &  Test Acc.  & 0.916(0.046)  &  \textbf{0.978(0.022)} &  & 0.950(0.039) & 0.948(0.022) & 0.958(0.020) \\
            \midrule   
            \multirow{2}{*}{Raisin}    & Train Acc. & 0.864(0.007)  &  0.863(0.004) & 0.862(0.007)  &  0.866(0.003) & 0.864(0.005)  & \textbf{0.867(0.001)} \\
                &  Test Acc.  & 0.843(0.010)   & 0.842(0.010)  & 0.843(0.012)  &  \textbf{0.847(0.004)} &  0.838(0.007) & 0.844(0.002) \\
            \bottomrule
            \end{tabular}
    \begin{tablenotes}\footnotesize
    \item[1] Uses Binomial likelihood (only two output classes allowed)
    \item[2] The DCC-Tree algorithm predicted similar class probabilities across each chain when applied to the Iris dataset, resulting in the same overall predicted class, and the small variance in the metrics as shown here. 
    \end{tablenotes}
    \end{threeparttable}
        }
\end{small}
\end{scshape}
\end{table}

Although DCC-Tree is not the best-performing method for any of the real-world datasets, it does perform similarly to the other HMC-based methods. One thing that is important to note is that in the majority of cases, the variance in the computed metrics is significantly less than in the remaining methods. This means that the DCC-Tree sampling routine is much more consistent than other methods in generating the required samples.

\section{Discussion}

We have proposed a novel sampling technique for Bayesian decision trees that incorporates the efficiency of HMC into the DCC framework to provide an efficient and fast sampling algorithm. Motivated by the work of \citet{zhou2020divide} in the context of probabilistic programming, the tree-sampling problem has been reduced into its constituents such that local inference can be applied, and then appropriately combined to give an overall estimate of the distribution. The efficacy of the sampling method developed in this paper has been demonstrated on a range of synthetic and real-world datasets and compared to existing methods. 

By associating a unique parameter vector to each tree topology, the DCC-Tree method successfully implements a more natural way of sampling from the posterior distribution. The space is divided based on the different tree topologies and local inference is conducted within each subspace, combining the resultant samples to recover the overall posterior. The method keeps track of each considered tree topology such that the burn-in period is only required once per unique structure reducing the per-proposal complexity of other HMC-based approaches.
This also enables the full benefits of the HMC sampling routine to be exploited, which is likely the reason behind the observed performance.

Comparing the performance on the synthetic datasets shows that the DCC-Tree algorithm gives the best testing performance out of all Bayesian decision tree methods (see Table \ref{tab:dcc-synth-metrics}), however, this is only just better than other HMC-based algorithms. Furthermore, the overfitting problem that is prevalent in other MCMC-based methods remains a non-issue here. This further motivates the use of HMC for the local inference routine and for exploring the posterior of decision trees in general. 

The performance results on the real-world datasets, Table \ref{tab:dcc-real-metrics}, show that the DCC-Tree method performs similarly to the other HMC-based methods, outperforming non-HMC methods on all but one dataset. On top of the performance, one major benefit of the DCC-Tree method is consistency. This is best shown by considering the standard deviation of the test performance in both Table~\ref{tab:dcc-synth-metrics} and Table~\ref{tab:dcc-real-metrics}. For the synthetic datasets, the standard deviation values are nearly an order of ten better than the next best method. When applied to real-world datasets, in nearly all cases, the standard deviation values are close to half that of any other method. 

Investigation of the DCC-Tree method has revealed that trees that include the correct tree as a subtree (i.e. additional nodes either beneath or above) have similar marginal likelihood values to that of the true underlying tree.  This is because the likelihood of the tree is a combination of the local likelihood at each leaf node. If the data is split in the same way (with some leaves identically distributed) then the overall likelihood is relatively similar. Intuitively, these trees are likely finding the same split in the data or are set to one of the local splitting extremities such that additional nodes are obsolete. However, it is possible to penalise larger trees through the prior on the tree structure, which has been implemented here. Therefore, it makes sense that the correct tree structures are given the highest marginal likelihood over all trees, as can be observed for the WU dataset in Figure \ref{fig:dcc-marg-llh-cgm-wu}. Despite this, the results shown in Figure \ref{fig:dcc-marg-llh-cgm-wu} for the CGM dataset show that this may not always be the case, as the inference scheme may not find the mode with the highest likelihood in the local posterior distribution.

One further benefit of the DCC-Tree method is that the significant burn-in overhead of the RJHMC-Tree method has been alleviated. However, what remains an issue with this sampling method, is that once a tree structure has been initialised, further sampling remains around the discovered area of the local posterior distribution. Although somewhat mitigated through the use of multiple chains, this problem remains, and is particularly evident when considering the marginal likelihood estimates for the CGM dataset in Figure \ref{fig:dcc-marg-llh-cgm-wu}. Possible extensions of this work could attempt to address this issue by defining something along the lines of a `restart' in which the same tree structure is burnt in again to see if any new modes are found.

Although it appears to improve on existing Bayesian decision tree methods, there are some shortcomings to the DCC-Tree algorithm. Just like with all other methods, new tree topologies are discovered in a random-walk-like fashion. Some improvement to this has been made by using the utility function to easily jump between the different structures, but this is only applicable to those trees that have been labelled `active'. New trees are still only proposed via random proposals local to the current tree. In addition to this, and as discussed previously, the estimated marginal likelihood of the true tree is similar to those that contain it as a subtree. As a result, the utility function does allocate significant exploration effort to these larger trees. Furthermore, the division by the number of times selected (see Equation \ref{eq:utility}) quickly averages out the small difference in the computed marginal likelihood. Nonetheless, these trees could be considered just as good as the true underlying tree if the only metric of interest was the predictive ability.

Overall, the DCC-Tree algorithm appears to have made a significant improvement in exploring the posterior distribution of the decision tree. By altering the underlying ideology, the approach has moved from one where the proposal scheme couples the tree structure and decision parameters to one where each distinct topology is uniquely associated with a parameter vector. Thus, each tree topology is considered individually, with the generated local samples combined to recover the overall posterior. 
This paper has demonstrated that considering the exploration of the Bayesian decision tree posterior in this way improved performance and consistency with respect to other inference methods.

\newpage

\bibliographystyle{abbrvnat}
\bibliography{main}

\begin{thebibliography}{23}
\providecommand{\natexlab}[1]{#1}
\providecommand{\url}[1]{\texttt{#1}}
\expandafter\ifx\csname urlstyle\endcsname\relax
  \providecommand{\doi}[1]{doi: #1}\else
  \providecommand{\doi}{doi: \begingroup \urlstyle{rm}\Url}\fi

\bibitem[Betancourt(2017)]{betancourt2017conceptual}
M.~Betancourt.
\newblock {A Conceptual Introduction to Hamiltonian Monte Carlo}.
\newblock \emph{arXiv preprint arXiv:1701.02434}, 2017.

\bibitem[Breiman et~al.(1984)Breiman, Friedman, Olshen, and
  Stone]{breiman1984classification}
L.~Breiman, J.~Friedman, R.~Olshen, and C.~Stone.
\newblock {Classification and regression trees. Wadsworth Int}.
\newblock \emph{Group}, 37\penalty0 (15):\penalty0 237--251, 1984.

\bibitem[Chipman et~al.(1998)Chipman, George, and
  McCulloch]{chipman1998bayesian}
H.~A. Chipman, E.~I. George, and R.~E. McCulloch.
\newblock {Bayesian CART Model Search}.
\newblock \emph{Journal of the American Statistical Association}, 93\penalty0
  (443):\penalty0 935--948, 1998.

\bibitem[Cochrane et~al.(2023)Cochrane, Wills, and
  Johnson]{cochrane2023rjhmctree}
J.~A. Cochrane, A.~G. Wills, and S.~J. Johnson.
\newblock {RJHMC-Tree for Exploration of the Bayesian Decision Tree Posterior}.
\newblock \emph{arXiv preprint arXiv:2312.01577}, 2023.

\bibitem[Denison et~al.(1998)Denison, Mallick, and Smith]{denison1998bayesian}
D.~G. Denison, B.~K. Mallick, and A.~F. Smith.
\newblock {A Bayesian CART algorithm}.
\newblock \emph{Biometrika}, 85\penalty0 (2):\penalty0 363--377, 1998.

\bibitem[Dua and Graff(2017)]{Dua:2019}
D.~Dua and C.~Graff.
\newblock {"{UCI} Machine Learning Repository"}, 2017.
\newblock URL \url{http://archive.ics.uci.edu/ml}.

\bibitem[Ghahramani(2015)]{ghahramani2015probabilistic}
Z.~Ghahramani.
\newblock Probabilistic machine learning and artificial intelligence.
\newblock \emph{Nature}, 521\penalty0 (7553):\penalty0 452--459, 2015.

\bibitem[Gramacy and Lee(2008)]{gramacy2008bayesian}
R.~B. Gramacy and H.~K.~H. Lee.
\newblock {Bayesian Treed Gaussian Process Models with an Application to
  Computer Modeling}.
\newblock \emph{Journal of the American Statistical Association}, 103\penalty0
  (483):\penalty0 1119--1130, 2008.

\bibitem[Hastings(1970)]{hastings1970monte}
W.~K. Hastings.
\newblock {Monte Carlo sampling methods using Markov chains and their
  applications}.
\newblock \emph{Biometrika}, 57\penalty0 (1):\penalty0 97--109, 1970.

\bibitem[Hoffman and Gelman(2014)]{hoffman2014no}
M.~D. Hoffman and A.~Gelman.
\newblock {The No-U-Turn Sampler: Adaptively Setting Path Lengths in
  Hamiltonian Monte Carlo.}
\newblock \emph{Journal of Machine Learning Research}, 15\penalty0
  (1):\penalty0 1593--1623, 2014.

\bibitem[Lakshminarayanan et~al.(2013)Lakshminarayanan, Roy, and
  Teh]{lakshminarayanan2013top}
B.~Lakshminarayanan, D.~Roy, and Y.~W. Teh.
\newblock {Top-down particle filtering for Bayesian decision trees}.
\newblock In \emph{International Conference on Machine Learning}, pages
  280--288. PMLR, 2013.

\bibitem[Linero and Yang(2018)]{linero2018bayesian}
A.~R. Linero and Y.~Yang.
\newblock Bayesian regression tree ensembles that adapt to smoothness and
  sparsity.
\newblock \emph{Journal of the Royal Statistical Society: Series B (Statistical
  Methodology)}, 80\penalty0 (5):\penalty0 1087--1110, 2018.

\bibitem[Llorente et~al.(2023)Llorente, Martino, Delgado, and
  Lopez-Santiago]{llorente2023marginal}
F.~Llorente, L.~Martino, D.~Delgado, and J.~Lopez-Santiago.
\newblock Marginal likelihood computation for model selection and hypothesis
  testing: An extensive review.
\newblock \emph{SIAM Review}, 65\penalty0 (1):\penalty0 3--58, 2023.

\bibitem[Metropolis et~al.(1953)Metropolis, Rosenbluth, Rosenbluth, Teller, and
  Teller]{metropolis1953equation}
N.~Metropolis, A.~W. Rosenbluth, M.~N. Rosenbluth, A.~H. Teller, and E.~Teller.
\newblock {Equation of State Calculations by Fast Computing Machines}.
\newblock \emph{The Journal of Chemical Physics}, 21\penalty0 (6):\penalty0
  1087--1092, 1953.

\bibitem[Neal et~al.(2011)]{neal2011mcmc}
R.~M. Neal et~al.
\newblock {MCMC using Hamiltonian dynamics}.
\newblock \emph{Handbook of Markov Chain Monte Carlo}, 2\penalty0 (11), 2011.

\bibitem[Pratola et~al.(2016)]{pratola2016efficient}
M.~T. Pratola et~al.
\newblock {Efficient Metropolis--Hastings Proposal Mechanisms for Bayesian
  Regression Tree Models}.
\newblock \emph{Bayesian Analysis}, 11\penalty0 (3):\penalty0 885--911, 2016.

\bibitem[Quinlan(1986)]{quinlan1986induction}
J.~R. Quinlan.
\newblock Induction of decision trees.
\newblock \emph{Machine Learning}, 1:\penalty0 81--106, 1986.

\bibitem[Quinlan(1993)]{quinlan1993c4}
J.~R. Quinlan.
\newblock \emph{C4.5: Programs for Machine Learning}.
\newblock Morgan Kaufmann, 1993.

\bibitem[Rainforth et~al.(2018)Rainforth, Zhou, Lu, Teh, Wood, Yang, and van~de
  Meent]{rainforth2018inference}
T.~Rainforth, Y.~Zhou, X.~Lu, Y.~W. Teh, F.~Wood, H.~Yang, and J.-W. van~de
  Meent.
\newblock Inference trees: Adaptive inference with exploration.
\newblock \emph{arXiv preprint arXiv:1806.09550}, 2018.

\bibitem[{Stan Development Team}(2019)]{stan2019}
{Stan Development Team}.
\newblock {Stan Modeling Language Users Guide and Reference Manual Version
  2.29}.
\newblock \emph{\url{https://mc-stan.org}}, 2019.

\bibitem[Taddy et~al.(2011)Taddy, Gramacy, and Polson]{taddy2011dynamic}
M.~A. Taddy, R.~B. Gramacy, and N.~G. Polson.
\newblock Dynamic trees for learning and design.
\newblock \emph{Journal of the American Statistical Association}, 106\penalty0
  (493):\penalty0 109--123, 2011.

\bibitem[Wu et~al.(2007)Wu, Tjelmeland, and West]{wu2007bayesian}
Y.~Wu, H.~Tjelmeland, and M.~West.
\newblock {Bayesian CART: Prior Specification and Posterior Simulation}.
\newblock \emph{Journal of Computational and Graphical Statistics}, 16\penalty0
  (1):\penalty0 44--66, 2007.

\bibitem[Zhou et~al.(2020)Zhou, Yang, Teh, and Rainforth]{zhou2020divide}
Y.~Zhou, H.~Yang, Y.~W. Teh, and T.~Rainforth.
\newblock Divide, conquer, and combine: a new inference strategy for
  probabilistic programs with stochastic support.
\newblock In \emph{International Conference on Machine Learning}, pages
  11534--11545. PMLR, 2020.

\end{thebibliography}

\newpage
\appendix

\section{Appendix}

\subsection{Validity of Method} \label{sec:app-valid}

\subsubsection{Problem Statement and Assumptions}\label{sec:dcc-assumpt}

An important prerequisite for the proof of correctness is ensuring that the division and recombination of the parameter space results in the correct overall posterior distribution. These assumptions and definitions required for this discussion will be presented now, starting with the parameter vector in Definition \ref{def:dcc-param-vec}.

\begin{definition}[DCC-Tree Parameter Vector]\label{def:dcc-param-vec}
    Let $\mathcal{M}$ be a discrete random variable whose value specifies a unique tree structure. Further, let the random variable $\Theta_m \in \bm{\Theta}_m \subset \mathbb{R}^{n_m}$ represent the associated parameters of each local tree subspace with appropriate dimension. Assume that there are $M$ topologies that span the entire parameter space such that $\mathcal{M} \in \{ 1,\dots,M\} $. The DCC-Tree parameter vector is then defined as,
\begin{equation}
    \Theta =  \left[ \begin{matrix} \Theta_1 & \Theta_2 & \cdots & \Theta_M \end{matrix} \right] \in \bm{\Theta},
\end{equation}
where
\begin{equation}
    \bm{\Theta} = \prod_{m=1}^M \bm{\Theta}_m \subset \mathbb{R}^{n_\Theta}, \quad n_\Theta = \sum_{m=1}^M n_m.
\end{equation}
\end{definition}

Careful consideration needs to be made as to the precise definition of the sample space. First, it is assumed that each subspace $\bm{\Theta}_i$ has compact support. Second, each parameter subspace is disjoint and covers the entire space,  i.e. $\bm{\Theta}_i \cap \bm{\Theta}_j = \emptyset$.  The overall parameter space includes both the parameter vector $\Theta$, which represents the continuous variables, and the discrete variable $\mathcal{M}$, as is shown in Definition \ref{def:dcc-space}.

\begin{definition}[DCC-Tree Parameter Space]\label{def:dcc-space}
    The DCC-Tree joint parameter-index vector $( \Theta, \mathcal{M} )$ is defined on the mixed discrete-continuous space,
    \begin{equation}
        \mathbb{S} := \mathbb{R}^{n_{\Theta}} \times \mathbb{Z}.
    \end{equation}
\end{definition}

Assumption \ref{asmp:dcc-model} defines how the output distribution can be simplified when given the tree subspace $m$.

\begin{assumption}\label{asmp:dcc-model}
    It is assumed that given the subtree under consideration, the output can be modelled as,
    \begin{equation}\
        p(y \mid x, \Theta, \mathcal{M} = m) \equiv p(y \mid x, \Theta_m).
    \end{equation}
    where $\mathcal{M}$ denotes the discrete random variable representing the tree topology under consideration. That is to say, given that the value of the discrete variable is $m$, the output distribution depends only on the corresponding random variable $\Theta_m$. 
\end{assumption}

Note that the remainder of this section will use the simplification in notation whereby when the discrete variable takes on a specific value then the discrete random variable will be dropped, i.e. instead of $\mathcal{M}=m$ is simplified to just $m$.

The following assumption defines an important property of the sample space once a specific subspace has been selected. In particular, it assumes that the space corresponding to the remaining parameter vector is uniform over the compact space, such that it integrates to one. This is necessary for the last theorem of this section, which will show that under this assumption, only local samples are required for the approximation of the posterior predictive distribution. 

\begin{assumption}\label{asmp:dcc-int}
    Let $\Theta_{\backslash m} \in \bm{\Theta}_{\backslash m}$ denote the component of the parameter vector defined by $\Theta \backslash \Theta_m$. 
    When considering subspace $m$, the distribution of $\Theta_{\backslash m}$ is assumed to be uniform over the compact space $\bm{\Theta}_{\backslash m}$. That is,
    \begin{equation}
        p(\Theta_{\backslash m}\mid m) = c_m
    \end{equation}
        where the value of $c_m$ is such that 
    \begin{equation}
        \int_{\bm{\Theta}_{\backslash m}} p(\Theta_{\backslash m}\mid m) d\Theta_{\backslash m} = 1.
    \end{equation}
\end{assumption}

Assumption \ref{asmp:dcc-int} enables simplification of the joint prior distribution across the continuous random variables $\Theta$ and the discrete random variable $\mathcal{M}$, as is discussed in the following lemma.

\begin{lemma}[Prior Distribution]\label{lem:dcc-prior}
    If the discrete random variable takes on the value $m$, the joint prior distribution $p(\Theta, m)$ can be expressed as,
    \begin{equation} \label{eq:dcc-dist-prior}
        p(\Theta, m) = c_m p(\Theta_m \mid m)p(m). 
    \end{equation}
    \proof
    Through the repeated application of conditional probability, the joint prior distribution can be written as,
    \begin{align}
        p(\Theta, m) &= p(\Theta \mid m)p(m) \\
        &= p(\Theta_m \mid m) p(\Theta_{\backslash m} \mid m) p(m) \\
        &= c_m p(\Theta_m \mid m) p(m)
    \end{align}
    where the second line is due to the independence of $\Theta_m$ and $\Theta_{\backslash m}$ and the third follows from Assumption \ref{asmp:dcc-int}. \qed
\end{lemma}

The purpose of this method is to approximate the posterior distribution such that expectations with respect to this distribution can be evaluated. Given a dataset $\mathcal{D} = \{ \mathbf{X}, \mathbf{Y} \}$, this  corresponds to generating samples from the joint posterior distribution $p(\Theta,\,  \mathcal{M} \mid \mathcal{D})$. The challenge with drawing samples from this joint posterior is that it is a mixed continuous-discrete distribution. However, once a particular subspace has been selected, inference on that subset of parameters is straightforward. The remainder of this section will discuss the connection between the posterior distribution $p(\Theta, m \mid \mathcal{D})$ and the distribution on the local parameters $p(\Theta_m \mid \mathcal{D}, m)$. This discussion begins by first introducing some notation that will simplify the later theorems. 

\begin{notation}\label{def:dcc-local-dist}
    Let the discrete random variable take on the value $m$. Then the following notation is defined with respect to the distribution $p(\Theta_m \mid \mathcal{D}, m)$,
    \begin{align}
        p_m(\Theta_m) & \triangleq p(\Theta_m \mid \mathcal{D}, m) \\
        Z_m & \triangleq p(\mathbf{Y} \mid m , \mathbf{X}).
    \end{align}
    Note that this implies the following relationship,
    \begin{equation} \label{eq:dcc-local-dist}
        p_m(\Theta_m) \, Z_m = p( \mathbf{Y} \mid \Theta_m, \mathbf{X})\, p(\Theta_m \mid m).
    \end{equation}
\end{notation}

The following theorem provides the relationship between the posterior distribution $p(\Theta,m \mid \mathcal{D})$ and the distribution on the local parameters $p(\Theta_m \mid \mathcal{D}, m)$.

\begin{theorem}[Posterior Distribution]\label{the:dcc-joint-post}
    Given a dataset $\mathcal{D}$, the posterior distribution can be expressed with respect to the local parameter distribution as,
    \begin{equation}
        p(\Theta,m \mid \mathcal{D}) = \bar{w}_m\, c_m \, p_m(\Theta_m).
    \end{equation}
    where
    \begin{equation}
        \bar{w}_m = \frac{Z_m p(m)}{\sum_{m=1}^M Z_m\, p(m)}
    \end{equation}
\proof
     The application of Bayes Theorem to the posterior distribution gives the following,
    \begin{align}
        p(\Theta, m \mid \mathcal{D}) &= \frac{p(\mathbf{Y} \mid \Theta, m, \mathbf{X})\, p(\Theta, m \mid \mathbf{X})}{p(\mathbf{Y} \mid \mathbf{X})},\\
        &= \frac{p(\mathbf{Y} \mid \Theta_m, \mathbf{X})\, p(\Theta_m \mid m)\, p(m)\, c_m}{p(\mathbf{Y} \mid \mathbf{X})}, \\
        &= \frac{p_m(\Theta_m)\, Z_m\, p(m)\, c_m}{p(\mathbf{Y} \mid \mathbf{X})}, \label{eq:dcc-joint-post-1}
    \end{align}
    where the second line follows from Equation \ref{eq:dcc-dist-prior}, and the third line from Equation \ref{eq:dcc-local-dist}. Using the law of total probability, the denominator can be expressed as,
    \begin{align}
    p(\mathbf{Y} \mid \mathbf{X}) &= \sum_{m=1}^M p(\mathbf{Y},m \mid \mathbf{X})\\
    &= \sum_{m=1}^M p(\mathbf{Y} \mid m, \mathbf{X})\, p(m)\\
    &= \sum_{m=1}^M Z_m\, p(m). 
    \end{align}
    Substitution into Equation \ref{eq:dcc-joint-post-1} gives the required expression,
    \begin{align}
        p(\Theta, m \mid \mathcal{D}) &= \frac{ Z_m\, p(m)}{\sum_{m=1}^M Z_m\, p(m)}\, c_m \,p_m(\Theta_m)\\
        &=  \bar{w}_m\, c_m \,p_m(\Theta_m),
    \end{align}
    with the normalised weights defined as,
    \begin{equation}
        \bar{w}_m = \frac{Z_m p(m)}{\sum_{m=1}^M Z_m\, p(m)}.
    \end{equation}
    \qed
\end{theorem}

Theorem \ref{the:dcc-joint-post} shows that the parameter space for the DCC-Tree algorithm can be thought of as a distribution where, once a subspace has been selected, only the corresponding subset of the overall parameter vector is important. The only concern is the presence of the constant $c_m$ which is difficult to compute. Fortunately, as will be shown in the following theorem, this is not of concern when evaluating expectation integrals due to the cancellation of the term.

\begin{theorem}[Estimation of Posterior Predictive Distribution]\label{the:dcc-int}
    Given a new datapoint $\{\mathbf{x}^*,y^*\}$, the posterior predictive distribution $p(y^* \mid \mathbf{x}^*, \mathcal{D}) $ can be approximated using samples from the joint posterior distribution $p(\Theta, m \mid\mathcal{D})$  as follows,
    \begin{equation}
        p(y^* \mid \mathbf{x}^*, \mathcal{D}) \approx \frac{1}{L} \sum_{i=1}^L p(y^* \mid \mathbf{x}^*, \Theta_{m^i}^i), \qquad (m^i, \Theta_{m^i}^i) \overset{\text{i.i.d.}}{\sim} \bar{w}_m\, p_m(\Theta_m).
    \end{equation}
    \proof
    Recall that the posterior predictive distribution is an expectation with respect to some distribution. In this case, this is the joint discrete-continuous distribution, with the expectation defined to be,
    \begin{equation}
        p(y^* \mid \mathbf{x}^*, \mathcal{D}) = \sum_{m=1}^M \int_{\bm{\Theta}} p(y^* \mid \mathbf{x}^*, \Theta, m, \mathcal{D})\, p(\Theta, m \mid \mathcal{D}) \text{d} \Theta       
    \end{equation}
    Using Assumption \ref{asmp:dcc-model}, Assumption \ref{asmp:dcc-int} and the result from Theorem \ref{the:dcc-joint-post}, this expression can be simplified in the following manner,
    \begin{align}
        p(y^* \mid \mathbf{x}^*, \mathcal{D}) &= \sum_{m=1}^M \int_{\bm{\Theta_m}} \int_{\bm{\Theta_{\backslash m}}} p(y^* \mid \mathbf{x}^*, \Theta, m, \mathcal{D})\, p(\Theta, m \mid \mathcal{D}) \text{d} \Theta_{\backslash m} \text{d} \Theta_m  \\
        &= \sum_{m=1}^M \int_{\bm{\Theta_m}} \int_{\bm{\Theta_{\backslash m}}} p(y^* \mid \mathbf{x}^*, \Theta_m)\, \bar{w}_m\, c_m \, p_m(\Theta_m) \text{d} \Theta_{\backslash m} \text{d} \Theta_m \\
        &= \sum_{m=1}^M \int_{\bm{\Theta_m}} p(y^* \mid \mathbf{x}^*, \Theta_m )\, \bar{w}_m\, p_m(\Theta_m) \text{d} \Theta_m 
    \end{align}    
    where it has been used that $y^*$ is independent of $\mathcal{D}$ given the parameter vector $\Theta$. The final expression shows that the posterior predictive distribution can be evaluated by considering only the weighted local distribution for each subspace $m$. This means that the integral can be approximated using Monte Carlo integration via
    \begin{align}
    p(y^* \mid \mathbf{x}^*, \mathcal{D}) &\approx \frac{1}{L} \sum_{i=1}^L \frac{p(y^* \mid \mathbf{x}^*, \Theta_{m^i}^i)\, \bar{w}_{m^i}\, p_{m^i}(\Theta_{m^i}^i)}{q(m^i,\Theta_{m^i}^i)}, \qquad (m^i, \Theta_{m^i}^i) \overset{\text{i.i.d.}}{\sim} q(m,\Theta_m)
    \end{align}
    where $q(\cdot)$ is a user-specified proposal distribution. In particular, this proposal can be taken as
    \begin{align}
    q(m,\Theta_m) &\triangleq \bar{w}_m\, p_m(\Theta_m)
    \end{align}
    which results in the following simplification
    \begin{align}
    p(y^* \mid \mathbf{x}^*, \mathcal{D}) &\approx \frac{1}{L} \sum_{i=1}^L p(y^* \mid \mathbf{x}^*, \Theta_{m^i}^i) \qquad (m^i, \Theta_{m^i}^i) \overset{\text{i.i.d.}}{\sim} \bar{w}_m\, p_m(\Theta_m),
    \end{align}
    giving the required expression. \qed 
\end{theorem}

It is clear from Theorem \ref{the:dcc-joint-post} that, when a particular subspace $\Theta_m$ is under consideration, the joint posterior simplifies down to the weighted local distribution. Further, Theorem \ref{the:dcc-int} shows that only samples from each parameter subspace $\Theta_m$ are required to generate the estimate of $p_m(\Theta_m)$ and therefore approximate integrals regarding the overall distribution. It was important to establish these concepts before continuing to the next section, which details the correctness of the method by considering the sampling method to be broken up into local and global components.

\subsubsection{Correctness of Method}

The correctness of the overall DCC-Tree sampling method relies on two parts: the validity of the local inference method to converge to the local target posterior distribution and the consistency of the local density approximations to recover the overall distribution. 
The method is based on the idea that the sample space can be split up and approximated as follows,
\begin{align}
    p(\Theta \mid \mathcal{D}) = \sum_{m=1}^M p(\Theta, m \mid \mathcal{D}) &= \sum_{m=1}^M  \frac{ c_m \,Z_m\, p(m)}{\sum_{m=1}^M Z_m\, p(m)} \, p(\Theta_m \mid m, \mathcal{D}) \\
    &\approx \sum_{m=1}^M  \frac{ c_m \, \hat{Z}_m\, p(m)}{\sum_{m=1}^M \hat{Z}_m\, p(m)} \, \hat{p}(\Theta_m \mid m, \mathcal{D}) \triangleq \hat{p}(\Theta \mid \mathcal{D}) \label{eq:dcc-dist-est}
\end{align}
where $\hat{Z}_m$ and $\hat{p}(\Theta \mid m, \mathcal{D}) $ are the approximations for $Z_m$ and $p(\Theta \mid m, \mathcal{D})$ respectively. Note that this shows that the distribution $ p(\Theta \mid \mathcal{D})$ can not be expressed directly as it contains the term $c_m$, but as shown in Theorem \ref{the:dcc-int} for the posterior predictive distribution, the samples generated from the local distribution $p(\Theta_m \mid m, \mathcal{D})$ can be used to evaluate expectation integrals with respect to this distribution. For the DCC-Tree algorithm, HMC is used to generate these local samples that can be used to provide these estimates.

HMC attempts to sample from a target distribution by using well-informed proposals that incorporate information about the likelihood distribution into the proposal scheme. For the DCC-Tree method, the target distribution is the distribution on the local parameters $p(\Theta_m \mid m, \mathcal{D})$. HMC draws samples from this distribution by using the relationship via Bayes' theorem as follows (note that the dependence on data on data $\mathcal{D}=\{\mathbf{X},\mathbf{Y}\}$ is explicitly included here):
\begin{equation}\label{eq:dcc-hmc}
    p(\Theta_m \mid m, \mathcal{D}) = \frac{p(\mathbf{Y} \mid \Theta_m, m, \mathbf{X}) p(\Theta_m \mid m ) }{p(\mathbf{Y} \mid m, \mathbf{X})} .
\end{equation}
Samples from the target distribution $p(\Theta_m \mid m, \mathcal{D})$ are generated in a way where only the unnormalised density (the numerator) is required due to the cancellation of terms in the accept/reject step of the algorithm \cite{metropolis1953equation,hastings1970monte}. Note that as a result of this, the denominator -- which is referred to as the marginal likelihood -- is also not required by the algorithm to produce the desired samples. 

The correctness of using HMC as the local inference scheme results from the validity of the sample algorithm itself, as is described in Lemma \ref{lem:HMC-validity}.

\begin{lemma}[Validity of Local Inference (HMC)]\label{lem:HMC-validity}
    Given the discrete random variable takes on the value $m$, the HMC algorithm generates samples from the local distribution $p(\Theta_m \mid m, \mathcal{D})$ as defined by Equation \ref{eq:dcc-hmc}.
    \proof
    As is standard with HMC sampling methods, the potential function is aligned with the negative-log likelihood of the unnormalised target distribution. For the DCC-Tree algorithm, this is defined to be,
    \begin{equation}
        U \triangleq - \log \left( p(\mathbf{Y} \mid \Theta_m, m, \mathbf{X}) \, p(\Theta_m \mid m )  \right)
    \end{equation}
    where $p(\mathbf{Y} \mid \Theta_m, m, \mathbf{X})$ is defined by either Equation \ref{eq:llh-class} for classification or Equation \ref{eq:llh-reg} for regression and $p(\Theta_m \mid m )$ by Equation \ref{eq:tree-prior}. The standard application of HMC under this potential energy definition then ensures that samples are generated from the normalised local distribution $p(\Theta_m \mid m, \mathcal{D})$. 
    \qed
\end{lemma}

The rest of this section focuses on the idea that, given the approximation techniques satisfy some assumptions, the estimated distribution can be recombined to give an estimate of the overall posterior distribution. Note that the rest of the section, including the next set of assumptions, follows the original proof provided in Appendix C of \citet{zhou2020divide} but where the notation has been changed to reflect that used in this paper.

\begin{assumption}\label{asmp:dcc-1}
    It is assumed that the total number M of tree subspaces $\Theta_m$ is finite.
\end{assumption}

\begin{assumption}\label{asmp:dcc-2}
    It is assumed that the number of iterations $T$ required to find all tree subspaces is almost surely finite. 
\end{assumption}

\begin{assumption}\label{asmp:dcc-3}
    Every tree subspace $m \in \{ 1,\dots, M \}$ has an associated local density estimate $\hat{p}(\Theta \mid m, \mathcal{D}) $ that converges weakly in the limit of large numbers to the true distribution $p(\Theta \mid m, \mathcal{D})$ corresponding to that subspace. Further, each tree subspace $m$ has a local marginal likelihood estimate $\hat{Z}_m$ that converges in probability to the true marginal likelihood of that subspace $Z_m$.
\end{assumption} 

With the above assumptions, Theorem \ref{the:dcc-correctness} proves the consistency of the DCC-Tree algorithm.

\begin{theorem}[Correctness of DCC-Tree Algorithm \cite{zhou2020divide}]\label{the:dcc-correctness}
    If Assumptions \ref{asmp:dcc-1} to \ref{asmp:dcc-3} hold, then the estimate of the posterior density defined by Equation \ref{eq:dcc-dist-est} generated by the DCC-Tree method converges weakly to the true distribution in the limit, that is,
    \begin{equation}
        \hat{p}(\Theta \mid \mathcal{D}) \to p(\Theta \mid \mathcal{D}) \qquad as \qquad T \to \infty.
    \end{equation}
    \proof
    The proof follows that presented in Theorem 1 in Appendix C of the supplementary material for \citet{zhou2020divide}, but is presented here with the notation related to this paper. For an arbitrary function $f$, the following result holds,
    \begin{align}
        \mathbb{E}_{\hat{p}(\Theta \mid \mathcal{D})} \left[ f(\Theta) \right] &= \int_{\bm{\Theta}}f(\Theta)\,   \hat{p}(\Theta \mid \mathcal{D}) \text{d} \Theta \\
        &= \int_{\bm{\Theta}}f(\Theta)\,  \sum_{m=1}^M  \frac{ c_m \, \hat{Z}_m\, p(m)}{\sum_{m=1}^M \hat{Z}_m\, p(m)} \, \hat{p}(\Theta \mid m, \mathcal{D}) \text{d} \Theta \\
        &= \sum_{m=1}^M \int_{\bm{\Theta}}f(\Theta)\, \frac{ c_m \, \hat{Z}_m\, p(m)}{\sum_{m=1}^M \hat{Z}_m\, p(m)} \, \hat{p}(\Theta \mid m, \mathcal{D}) \text{d} \Theta \\
        &= \sum_{m=1}^M \int_{\bm{\Theta}}f(\Theta)\, \frac{ c_m \, Z_m\, p(m)}{\sum_{m=1}^M Z_m\, p(m)} \, p(\Theta \mid m, \mathcal{D}) \text{d} \Theta \label{eq:slutsky}\\
        &= \int_{\bm{\Theta}}f(\Theta)\, \sum_{m=1}^M  \frac{ c_m \, Z_m\, p(m)}{\sum_{m=1}^M Z_m\, p(m)} \, p(\Theta \mid m, \mathcal{D}) \text{d} \Theta \\
        &= \int_{\bm{\Theta}}f(\Theta)\, p(\Theta \mid \mathcal{D}) \text{d} \Theta \\
         &=  \mathbb{E}_{p(\Theta \mid \mathcal{D})}\left[ f(\Theta) \right] 
    \end{align} 
    where Equation \ref{eq:slutsky} is due to Slutsky's theorem.  
    Therefore, the expectation of the function $f$ with respect to the approximated posterior distribution $\hat{p}(\Theta\mid \mathcal{D})$ converges to the expectation under the true distribution $p(\Theta\mid \mathcal{D})$. As $f$ was defined to be an arbitrary function then this result holds true in general, as required.     \qed
\end{theorem}

The next section will discuss the practical implementation of the DCC-Tree algorithm, and in particular, the methods for estimating the local distribution $\hat{p}(\Theta \mid m,\, \mathcal{D}) $ and the marginal likelihood $\hat{Z}_m$ for each subspace $m$.

\subsection{DCC-Tree Implementation Details}
Exact details on how certain calculations have been implemented are presented in this section. 

\subsubsection{Log-Marginal Likelihood Calculation}\label{sec:app-log-marg}
From Section \ref{sec:marg-llh-calc}, the marginal likelihood estimate for tree $\T_m$ is calculated via
\begin{equation*}
    \hat{Z}_m = \frac{1}{N_TN_cN_M} \sum_{i=1}^{N_T} \sum_{j=1}^{N_c} \sum_{k=1}^{N_M} \tilde{w}_{i,j,k}^{(m)}, \quad  \tilde{w}_{i,j,k}^{(m)} = \frac{\tilde{\pi}_m(\xi^{(i,j,k)})}{\Phi_m(\xi^{(i,j,k)})}
\end{equation*}
where  $ \pi_m(\xi^{(i,j,k)})$ is the posterior distribution on parameters and $\xi^{(i,j,k)}$ are the $k=1,\dots,N_M$ new parameter samples at iteration $i$ of chain $j$. The term $\Phi_m(\xi^{(i,j,k)})$ is computed using either the basic or spatial formulation,
\begin{equation*}
    \Phi_m(\xi^{(i,j,k)}) = q_{i,j,m}(\xi^{(i,j,k)} \mid \nu_m^{(i,j)},\Sigma), \text{ or, } \Phi_m(\xi^{(i,j,k)}) = \frac{1}{N_T} \sum_{n=1}^{N_T} q_{n,j,m}(\xi^{(i,j,k)} \mid \nu_m^{(n,j)},\Sigma)
\end{equation*}
These calculations are converted to log space as follows. The estimated log marginal likelihood is computed via 
\begin{equation}
    \log \hat{Z}_m = \text{logsumexp}\left(\log \tilde{w}_{i,j,k}^{(m)}\right) - \log(N_TN_cN_M)
\end{equation}
where 
\begin{equation}
    \log \tilde{w}_{i,j,k}^{(m)} = \log \tilde{\pi}_m\left(\xi^{(i,j,k)}\right) - \log \Phi_m\left(\xi^{(i,j,k)}\right).
\end{equation}
Access to $\log \pi_m\left(\xi^{(i,j,k)}\right)$ is available directly throughout the simulation. The term  $\log \Phi_m\left(\xi^{(i,j,k)}\right)$ is calculated either as
\begin{equation}
    \log \Phi_m\left(\xi^{(i,j,k)}\right) = \text{logsumexp}\left(\log q_{i,j,m}(\xi^{(i,j,k)})\right),
\end{equation} 
for the basic expression, or as the spatial version,
\begin{equation}
    \log \Phi_m\left(\xi^{(i,j,k)}\right) = \text{logsumexp}\left(\log q_{n,j,m}(\xi^{(i,j,k)})\right) - \log(N_T),
\end{equation} 
where the proposal distribution $q$ are initialised in log-form as required (see Section \ref{sec:marg-llh-calc} for details). 

Lastly, the log-marginal likelihood does not need to be recalculated at each additional visit to the tree. Instead, it can be updated via,
\begin{equation*}
    \log \hat{Z}_m^{(t)} = \text{logsumexp} \left( \left[\log \hat{Z}_m^{(t-1)}+\log((t-1)N_sN_cN_M), \log w_{i,j,k}^{(t-1)} \right] \right) - \log(tN_sN_cN_M)
\end{equation*}
where $t\in \mathbb{N}$ denotes the visit number to that tree.

\subsubsection{Calculation of Exploitation Term}\label{sec:app-exploit}

Through the simulation, the terms $\log(\hat{Z}_m)$ and $\log(\sigma_m^2)$ for each tree $\T_m$ are recorded. However, the value of these can be large, resulting in either 0 or an infinite value when taking the exponent and the loss of any relative information. Therefore, a numerically stable way is required to approximate the calculation of the exploitation term, which is computed as follows,
\begin{equation}\label{eq:taum}
    \frac{\hat{\tau}_m}{\max_m \{ \hat{\tau}_m \}}, \quad \hat{\tau}_m = \sqrt{\hat{Z}_m^2 + (1+\kappa)\sigma_m^2}.
\end{equation}
Only access to $\log(\hat{Z}_m)$ and $\log(\sigma_m^2)$ is provided as the method progresses. As such, the above expression for $\hat{\tau}_m$ is rewritten as
\begin{equation}\label{eq:taum2}
    \sqrt{\hat{Z}_m^2 + (1+\kappa)\sigma_m^2} = \sqrt{\exp \left( 2\log \hat{Z}_m \right) + (1+\kappa)\exp \left( \log \sigma_m^2 \right)}.
\end{equation}
Taking $A_m = \max \left\{ 2\log(\hat{Z}_m), \log(\sigma_m^2) \right\} $, the following transforms can be used to alter the above expression,
\begin{align*}
    \widetilde{2\log(\hat{Z}_m)} &= 2\log(\hat{Z}_m) - A_m  && \widetilde{\log(\sigma_m^2)} = \log(\sigma_m^2) - A_m
\end{align*}
where $A_m$ is now large. This changes the expression in Equation \ref{eq:taum2} to
\begin{align}
    \sqrt{\hat{Z}_m^2 + (1+\kappa)\sigma_m^2} &=\sqrt{\exp \left(A_m + \widetilde{ 2\log \hat{Z}_m} \right) + (1+\kappa)\exp \left( A_m + \widetilde{\log \sigma_m^2} \right)} \\
    &= \underbrace{\sqrt{\exp(A_m)}}_{\text{unstable}}\cdot \underbrace{\sqrt{\exp \left(\widetilde{ 2\log \hat{Z}_m} \right) + (1+\kappa)\exp \left(\widetilde{\log \sigma_m^2} \right)}}_{\text{can be calculated}}
\end{align}
Since both the square root and exponential functions are monotonically increasing, the ratio in Equation \ref{eq:taum} can be computed by evaluating the relative expression between tree topologies. To compute this ratio, assume that there are two tree topologies with subscripts $p$ and $q$ and assume without loss of generality that $q$ corresponds to the value for which $\hat{\tau}_m$ is maximum. The ratio for topology $p$ is then given by,
\begin{equation}
    \frac{\hat{\tau}_p}{\max_m \{ \hat{\tau}_m \}} = \sqrt{\exp(A_p-A_q)}\frac{\sqrt{\exp \left(\widetilde{ 2\log \hat{Z}_p} \right) + (1+\kappa)\exp \left(\widetilde{\log \sigma_p^2} \right)}}{\sqrt{\exp \left(\widetilde{ 2\log \hat{Z}_q} \right) + (1+\kappa)\exp \left(\widetilde{\log \sigma_q^2} \right)}}
\end{equation}
where all terms are now numerically stable to calculate. This method also handles cases when either the log-marginal likelihood or log-variance is significantly larger than the other.

\subsubsection{Calculation of Exploration Term}

The exploration term $\hat{\rho}_m$ as first described in \citet{rainforth2018inference} is adopted here. This term is used to help define the possible improvement in the marginal likelihood estimation with additional inference samples. The expression relates to the log-weights of the current set of samples as follows,
\begin{equation}
    \hat{\rho}_m \triangleq P\left( \hat{w}_m(T_a) > w_{th} \right) \approx 1 - \Psi_m(\log w_{th})^{T_a},
\end{equation}
and defines the probability that at least one new sample in a ``look-ahead'' horizon would have a log-weight greater than some threshold weight $w_{th}$. The function $\Psi_m$ is taken to be the cumulative distribution function for the normal distribution with mean and variance based on the current samples for tree $\T_m$. The value of the threshold weight $w_{th}$ is taken to be the maximum weight of the samples discovered so far across all active trees. The variable $T_a$ is a hyperparameter that represents the number of ``look-ahead'' samples in the horizon. As per \citet{rainforth2018inference}, a default value of $T_a=1000$ is used for the DCC-Tree algorithm.

\subsection{True Tree Definitions for Synthetic Datasets} \label{sec:app-true-trees}

\begin{figure}[ht]
    \begin{center}
    \begin{subfigure}[t]{.49\linewidth}
        \begin{center}
            \resizebox{0.5\textwidth}{!}{
                \begin{tikzpicture}
                    \node [internal]                                    (n0)    {$x_1 \leq 0.5$};
                    \node [internal, below=1cm of n0, xshift=-1cm]      (n1)    {$x_2 \leq 0.5$};
                    \node [leaf, below=1cm of n1, xshift=-1cm]          (l1)    {$\eta_{\ell_1}$};
                    \node [leaf, below=1cm of n1, xshift=1cm]           (l2)    {$\eta_{\ell_2}$};
                    \node [leaf, below=1cm of n0, xshift=1cm]           (l3)    {$\eta_{\ell_3}$};

                    \path [line] (n0) -- (n1);
                    \path [line] (n1) -- (l1);
                    \path [line] (n1) -- (l2);
                    \path [line] (n0) -- (l3);
                \end{tikzpicture}
            }
        \end{center}
    \end{subfigure}
    \begin{subfigure}[t]{.49\linewidth}
        \begin{center}
            \resizebox{0.5\textwidth}{!}{
                \begin{tikzpicture}
                    \node [internal]                                    (n0)    {$x_3 \leq 0.5$};
                    \node [internal, below=1cm of n0, xshift=1cm]       (n1)    {$x_2 \leq 0.5$};
                    \node [leaf, below=1cm of n1, xshift=-1cm]          (l1)    {$\eta_{\ell_1}$};
                    \node [leaf, below=1cm of n1, xshift=1cm]           (l2)    {$\eta_{\ell_2}$};
                    \node [leaf, below=1cm of n0, xshift=-1cm]           (l3)    {$\eta_{\ell_3}$};

                    \path [line] (n0) -- (n1);
                    \path [line] (n1) -- (l1);
                    \path [line] (n1) -- (l2);
                    \path [line] (n0) -- (l3);
                \end{tikzpicture}
            }
        \end{center}
    \end{subfigure}
    \caption{The two trees consistent with the synthetic dataset defined in \citet{wu2007bayesian}. }
    \label{fig:wu-tree}
\end{center}
\end{figure}
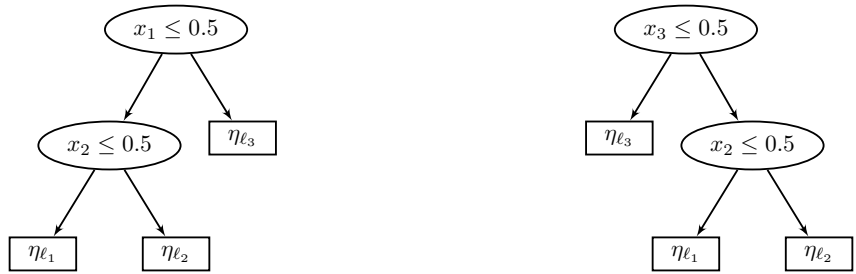

The synthetic dataset proposed by \citet{wu2007bayesian} is defined such that the input space is given by,
\begin{equation*}
    \bm{x}_i = 
    \begin{cases}
        (x_1, x_2, x_3) & \text{where } x_1 \sim \mathcal{U}_{[0.1,0.4]}, x_2 \sim \mathcal{U}_{[0.1,0.4]}, x_3 \sim \mathcal{U}_{[0.6,0.9]}, \text{ for } i = 1, \dots, 100,\\
        (x_1, x_2, x_3) & \text{where } x_1 \sim \mathcal{U}_{[0.1,0.4]}, x_2 \sim \mathcal{U}_{[0.6,0.9]}, x_3 \sim \mathcal{U}_{[0.6,0.9]}, \text{ for } i = 101, \dots, 200,\\
        (x_1, x_2, x_3) & \text{where } x_1 \sim \mathcal{U}_{[0.6,0.9]}, x_2 \sim \mathcal{U}_{[0.1,0.9]}, x_3 \sim \mathcal{U}_{[0.1,0.4]}, \text{ for } i = 201, \dots, 300,
    \end{cases}
    \\
\end{equation*}
where $\mathcal{U}_{[a,b]}$ represents the uniform distribution over the interval $[a,b]$. The corresponding outputs are then defined to be,
\begin{equation*}
    y =  
    \begin{cases}
        1 + \mathcal{N}(0,0.25) & \text{if $x_1 \leq 0.5$ and $x_2 \leq 0.5$},\\
        3 + \mathcal{N}(0,0.25) & \text{if $x_1 \leq 0.5$ and $x_2 > 0.5$},\\
        5 + \mathcal{N}(0,0.25) & \text{if $x_1 > 0.5$},
    \end{cases}
\end{equation*}
with the two trees that are consistent with the data (equally as likely) shown in Figure \ref{fig:wu-tree}.

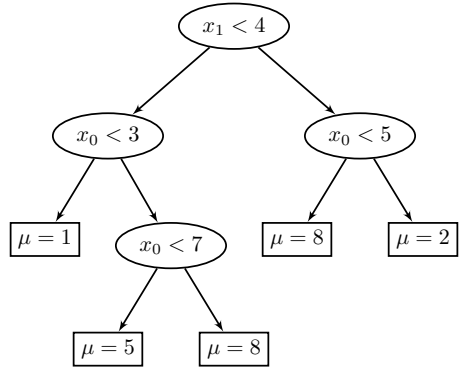
\begin{figure}[ht]
    \begin{center}
        \resizebox{0.4\textwidth}{!}{
        \begin{tikzpicture}
            \node [internal]                                        (n0)    {$x_1 < 4$};
            \node [internal, below=1cm of n0, xshift=-2cm]      (n1)    {$x_0 < 3$};
            \node [internal, below=1cm of n0, xshift=2cm]       (n2)    {$x_0 < 5$};
            \node [leaf, below=1cm of n1, xshift=-1cm]        (n3)    {$\mu=1$};
            \node [internal, below=1cm of n1, xshift=1cm]            (l3)    {$x_0 < 7$};
            \node [leaf, below=1cm of n2, xshift=-1cm]             (l4)    {$\mu=8$};
            \node [leaf, below=1cm of n2, xshift=1cm]             (l5)    {$\mu=2$};
            \node [leaf, below=1cm of l3, xshift=-1cm]             (l6)    {$\mu=5$};
            \node [leaf, below=1cm of l3, xshift=1cm]             (l7)    {$\mu=8$};
            \path [line] (n0) -- (n1);
            \path [line] (n0) -- (n2);
            \path [line] (n1) -- (n3);
            \path [line] (n1) -- (l3);
            \path [line] (n2) -- (l4);
            \path [line] (n2) -- (l5);
            \path [line] (l3) -- (l6);
            \path [line] (l3) -- (l7);
        \end{tikzpicture}
        }
        \caption{True tree definition used to generate the synthetic dataset adapted from \citet{chipman1998bayesian}. }
        \label{fig:cgm-tree}
    \end{center}
\end{figure}

The synthetic dataset of \citet{chipman1998bayesian}  was adapted in the same manner as \citet{cochrane2023rjhmctree}. This differs from the original version by making the first split defined by a continuous input not categorical. All inputs are numerical with $n_x = 2$ and simulated via the tree structure shown in Figure \ref{fig:cgm-tree} to give $N_\textsc{train} = 800$, $N_\textsc{test} = 800$ datapoints. The variance is assumed constant across nodes with value $\sigma^2 = 0.2^2$.

\subsection{Hyperparameters Optimisation}\label{sec:app-hyp}

Hyperparameters used for the simulation of Bayesian tree methods were taken from \citet{cochrane2023rjhmctree}. We provide analysis of an additional dataset in this paper - the synthetic dataset from \citet{wu2007bayesian}. Table \ref{tab:grid-search} provides the values used for the grid search, with Table \ref{tab:hyperparams} detailing the final hyperparameters used for each of the other Bayesian decision tree methods. Note that the SMC method is not included as its implementation is not applicable to regression datasets.

\begin{table}[h]
\centering
    \caption{Grid search values considered in hyperparameter optimisation 
 via 5-fold cross-validation for the synthetic dataset of \citet{wu2007bayesian}.}\label{tab:grid-search} 
\small\scshape
\begin{tabular}{cp{8cm}}
    \toprule
    Method  & Hyperparameters \\
    \midrule
    \multirow{2}{*}{WU}  & $\alpha = [0.5:0.5:4.0]$, $\beta = [0.5:0.5:4.0]$, $\mu_0 = [0:0.5:2]$, $n=[0.5:0.5:1.5]$, $\lambda = [8:1:10]$, $p = [0.3:0.2:0.7]$  \\
\midrule
\multirow{3}{*}{CGM}   & $\alpha = [0.5:0.5:4.0]$, $\beta = [0.5:0.5:2.5]$, $\mu_0 = [0:0.5:2]$, $n=[0.3:0.1:0.6]$, $\alpha_\textsc{split} = [0.45:0.25:0.95]$, $\beta_\textsc{split} = [1.0:0.5:2.5]$ \\
\bottomrule
\end{tabular}
\end{table}

\begin{table}[H]
    \caption{Final hyperparameters for the other Bayesian decision tree methods for the synthetic dataset of \citet{wu2007bayesian}}\label{tab:hyperparams}  
    \begin{center}
    \small\scshape
    \begin{tabular}{cp{9cm}}
    \toprule
    Method & Hyperparameters  \\
\midrule
    RJHMC  & $h_\textsc{init}=0.025$, $h_\textsc{final}=0.025$,  $\alpha_\textsc{split} = 0.45$, $\beta_\textsc{split} = 2.5$   \\
\midrule
WU & $\alpha = 4.0$, $\beta = 4.0$, $\mu_0 = 1.0$, $n=1.0$, $\lambda = 10.0$, $p = 0.7$ \\
\midrule     
CGM   & $\alpha = 3.5 $, $\beta = 0.5$, $\mu_0 = 1.0$, $n=0.5$, $\alpha_\textsc{split} = 0.95$, $\beta_\textsc{split} = 1.0$ \\
    \bottomrule
\end{tabular}
\end{center}
\end{table}

The final set of hyperparameters used for the DCC-Tree method simulations is provided in Table~\ref{tab:hyperparams-dcc}. Note that hyperparameters not listed are default values.

\begin{table}[H]
    \caption{Hyperparameters used for the DCC-Tree method for different datasets.}\label{tab:hyperparams-dcc}  
    \begin{center}
    \small\scshape
    \begin{tabular}{cp{4cm}}
    \toprule
     Dataset & Hyperparameters  \\
    \midrule
    bcw & $h_\textsc{init}=0.1$, $h_\textsc{final}=0.025$   \\
    cgm & $h_\textsc{init}=0.01$, $h_\textsc{final}=0.001$   \\
    iris & $h_\textsc{init}=0.01$, $h_\textsc{final}=0.01$   \\
    raisin & $h_\textsc{init}=0.05$, $h_\textsc{final}=0.001$   \\
    wine & $h_\textsc{init}=0.025$, $h_\textsc{final}=0.025$   \\
    wu & $h_\textsc{init}=0.5$, $h_\textsc{final}=0.025$   \\
    \bottomrule
\end{tabular}
\end{center}
\end{table}

\newpage

\end{document}